\documentclass[11pt]{article}

\usepackage[preprint]{acl}

\usepackage{times}
\usepackage{latexsym}
\usepackage{amsmath}
\usepackage{adjustbox}
\usepackage{booktabs}
\usepackage[T1]{fontenc}

\usepackage[utf8]{inputenc}

\usepackage{microtype}

\usepackage{graphicx}

\title{Inside VLM Chart Reading: Tracing Value Reading from Vertical Bar Charts Across Space and Depth}

\author{
Tianhao Niu
\quad
\textbf{Qingfu Zhu}
\quad
\textbf{Wanxiang Che} \\
 Research Center for Social Computing and Interactive Robotics \\
 Harbin Institute of Technology, China \\
}

\usepackage{hyperref}
\usepackage{cleveref}

\usepackage{ragged2e}

\usepackage{array}
\usepackage{tabularx}
\usepackage{makecell}
\usepackage{multirow}
\usepackage{listings}
\usepackage{xcolor}

\newcommand{\qwen}{Qwen2.5-VL}
\newcommand{\intern}{InternVL3.5}
\newcommand{\diffroi}{\textsc{Difference}}
\newcommand{\introi}{\textsc{Intersection}}
\newcommand{\unionroi}{\textsc{Union}}

\lstdefinestyle{barcompact}{
  language=Python,
  basicstyle=\ttfamily\scriptsize,
  keywordstyle=\bfseries,
  commentstyle=\itshape\color{gray},
  columns=fullflexible,
  keepspaces=true,
  showstringspaces=false,
  breaklines=true,
  breakatwhitespace=true,
  breakindent=0pt,
  frame=single,
  framerule=0.3pt,
  framesep=2pt,
  xleftmargin=1pt,
  xrightmargin=1pt,
  aboveskip=2pt,
  belowskip=2pt,
  tabsize=2
}
\usepackage{booktabs} 
\usepackage{tabularx} 
\usepackage{multirow}

\begin{document}
\maketitle

\begin{abstract}
Vision--language models (VLMs) can answer chart questions accurately, but output accuracy alone provides limited insight into the internal representations supporting exact value reading. We study vertical-bar value reading with controlled counterfactual activation patching in \qwen-7B-Instruct and \intern-8B. The study connects three analyses: 
(1) The single-factor analysis shows that, under one-tick bar-height counterfactuals, patching states selected by the changed-height region produces higher restoration than patching states selected by the invariant bar body. Across language-model depth, restoration from legend- and series-related positions declines earlier than restoration from bar-geometry and axis-scale positions.
(2)  We identify middle-layer prompt-series states can restore answer preference under legend swaps. Resetting these states selectively weakens restoration from an earlier visual-legend patch, revealing their causal role in the downstream effects of an earlier visual-legend intervention.
(3) In the factorial analysis, both models can use geometry and scale states from separate donors to favor the combined target. Together, these results provide causal evidence about the internal states supporting exact bar-value reading. \footnote{Our code and data will be released upon acceptance.}
\end{abstract}

\section{Introduction}
\label{sec:introduction}

Charts express numerical information through text, color, position, and geometric extent. Recent vision--language models (VLMs)~\citep{bai2025qwen25vltechnicalreport,wang2025internvl35advancingopensourcemultimodal,vteam2026glm5vturbonativefoundationmodel} perform well on chart question answering benchmarks~\citep{masry-etal-2022-chartqa,xu2024chartbenchbenchmarkcomplexvisual,wang2024charxivchartinggapsrealistic}. However, but output 
accuracy alone does not explain the internal states when conducting chart understanding.

To this end, we focus on reading the exact integer value of a specified vertical bar. The prompt identifies a series and category; the legend and x-axis labels connect them to a bar; the bar top gives a visual magnitude; and the y-axis maps that magnitude to a number. We ask three questions: (1) Where can states associated with different chart factors causally restore the original answer preference?   (2) When local restoration from visual-legend positions declines, can other states still restore answer preference under legend swaps, and do these states causally contribute to the downstream effect of an earlier visual-legend patch?    (3) Can VLMs combine bar-height and y-axis-scale internal states from different charts to recover the value implied by their combination?

We use controlled counterfactual activation patching to answer these questions. Paired charts or prompts differ in one or two known factor, and selected activation states are transferred between their forward passes. We score complete answer sequences and measure normalized restoration of the relevant answer preference. We evaluate Qwen2.5-VL-7B-Instruct and InternVL3.5-8B on matched examples.

\textbf{To answer the first question, we conduct  \textbf{single-factor analysis}.} The analysis separately changes axis scale, bar height, legend mapping, x-axis labels, and prompt targets. For bar height, we compare the full affected bar, the changed region around its top, and the unchanged body using the same chart pairs. 
Under one-tick bar-height counterfactuals, patching states selected by the changed-height region produces substantially higher restoration than patching states selected by the invariant bar body. This difference persists when the changed-height ROI contains no more selected tokens than the invariant-body ROI.  Across depth, restoration from legend- and series-related positions declines earlier than restoration from bar-geometry and axis-scale positions.

\textbf{The earlier decline in restoration at visual legend regions motivates the second question.} We first examine which token positions can restore the answer preference under a legend swap. By patching different positions across layers, we find that prompt-series tokens in the middle layers show clear restoration effects, much stronger than the prompt-label controls. However, this does not tell us whether these states contribute to the effect of an earlier visual-legend patch. To test this, we first restore the visual-legend states at a layer selected using pilot data, and then reset the prompt-series states at a deeper layer to their counterfactual values. This reset weakens the effect of the visual-legend patch, while resets at the control positions have much smaller effects. These results suggest  their causal role in the downstream effects of an earlier visual-legend intervention.

\textbf{To answer the third question, we conduct  \textbf{factorial recombination analysis}.} 
The analysis tests whether bar geometry and axis scale can be combined across forward passes. In each $2\times2$ family, a geometry donor changes only target-bar height, a scale donor changes only axis scale, and a diagonal chart contains both changes. Joint-cross patches geometry and scale states from the two separate donors into one receiver computation; joint-coherent takes both states from the diagonal. InternVL changes little between the two joint conditions, whereas Qwen gives lower restoration when the states come from separate donors. Thus, downstream computation can use separately obtained geometry and scale states together.

Our contributions are threefold\footnote{Our mechanistic analyses focus on successful value-reading cases, where both models correctly answer all relevant unpatched inputs (both endpoints of a counterfactual pair or all four charts in a factorial family). Failure cases are outside the scope of this study.}:
\begin{itemize}
    \item \textbf{Single-factor localization.} We separate the local effects of x-axis label, bar height, axis scale, and prompt information, showing stronger value-specific restoration at the changed bar top and earlier loss of casual restoration of legend- and series-related states.

    \item \textbf{Causal role of prompt-series states.} We identify middle-layer prompt-series states with restoration effects under legend-swap counterfactuals and show that resetting these states selectively reduces the effect of an earlier visual-legend intervention which suggest  their causal role in the downstream effects of an earlier visual-legend intervention.

    \item \textbf{Cross-factor combination.} A factorial intervention shows that both models can jointly use bar-geometry and axis-scale states taken from separate forward passes, while differing in sensitivity to donor context.
\end{itemize}

\section{Related work}
\paragraph{Chart understanding and diagnostic evaluation.}
Chart question answering has progressed from controlled benchmarks such as FigureQA~\citep{kahou2018figureqaannotatedfiguredataset}, PlotQA~\citep{methani2020plotqareasoningscientificplots}, and ChartQA~\citep{masry-etal-2022-chartqa} to more diverse evaluations including ChartQAPro~\citep{masry2025chartqaprodiversechallengingbenchmark}, CharXiv~\citep{wang2024charxivchartinggapsrealistic}, ChartBench~\citep{xu2024chartbenchbenchmarkcomplexvisual}, and EncQA~\citep{Mukherjee_2026}, with broader studies further documenting the limitations of general-purpose VLMs on chart reasoning. Chart-specialized models such as DePlot~\citep{liu2023deplotoneshotvisuallanguage}, MatCha~\citep{liu2023matchaenhancingvisuallanguage}, and UniChart~\citep{masry2023unichartuniversalvisionlanguagepretrained} improve visual extraction and numerical reasoning, but output-level evaluation alone cannot determine how model uses the series labels, category labels, bar geometry and axis-scale internally.

Most closely related, FUGU~\citep{tartaglini2025diagnosingbottlenecksdatavisualization} uses synthetic scatter plots, vision-encoder activation patching, and linear probes to diagnose coordinate-extraction and vision–language handoff bottlenecks. In contrast, we study successful exact value reading in vertical bar charts using more fine-grained component-wise image and prompt counterfactuals with geometry-aligned patching across space and depth.

\paragraph{Activation patching and multimodal interpretability.}
Causal tracing and activation patching replace hidden states between
paired inputs and measure the resulting change in model behavior
\citep{meng2023locatingeditingfactualassociations}. These interventions
support circuit localization and comparisons of information
availability across depth
\citep{NEURIPS2023_34e1dbe9}. However, conclusions can
depend strongly on how the corrupted input is constructed, which
output metric is used, and whether interventions target neurons,
tokens, layers, or larger components
\citep{zhang2024bestpracticesactivationpatching}. 

Mechanistic analysis has increasingly been extended to multimodal
models. Early work adapts causal tracing to BLIP for image-conditioned
generation \citep{palit2023visionlanguagemechanisticinterpretabilitycausal}. NOTICE replaces
unstructured Gaussian corruption with semantic image perturbations
and aligned text-token replacements, enabling more interpretable
causal mediation experiments in VLMs
\citep{golovanevsky2025vlmsnoticemechanisticinterpretability}. Recent work also patches broad sets
of image, query, and output-token states to distinguish direct from
text-mediated routes of visual information
\citep{salazar2026pathwaysvisualinformationflow}.

We build on these principles with chart-program-derived
counterfactuals and geometry-derived ROIs aligned to each model's
visual token grid. Original states are inserted into a
counterfactual computation whose correct numerical answer has
changed, and restoration is measured using the complete answer
sequence rather than a single output token.

\section{Preliminary}
\subsection{Task and Controlled Charts Value Reading Data Generation}

\paragraph{Task Definition} Each example contains a vertical bar chart and a query of the form: \emph{Read the exact y-axis value of the ``SERIES'' bar at ``LABEL''. Return only an integer.}

\paragraph{Chart Value Reading Data Generation} Series and category names are semantically meaningless pseuwords. Charts contain 1--3 series and 4--6 categories; the y-axis begins at zero, with major tick steps in $\{5,10\}$. y-max$\leq$60. Layout slots are fixed and grids, minor ticks, hatching, and decorative elements are removed. The same chart semantics are rendered at $1344{\times}1344$ for \qwen{} and $448{\times}448$ for \intern{}. The code template for each synthesized chart is shown in Figure~\ref{fig:bar-generation-template} and related parameters are randomly sampled. We synthesize 500 charts in total. For each chart and each bar in chart, we construct a value reading sample using the template in task definition and this result in 5932 samples. Detailed distribution are shown in appendix~\ref{sec:chartvaluereadingdatagendetail}.

\subsection{Counterfactual Chart Generation}

For the 5,932 value-reading queries, we construct controlled image and prompt counterfactuals. Image counterfactuals modify axis scale, bar height, legend mapping, or x-axis labels, while prompt counterfactuals replace the target category expression (Label), series expression (Series), or both (Label + series). For all prompt counterfactuals, replacement expressions are constrained to have the same token counts and occupy the same token positions as the original expressions, ensuring position-aligned activation patching. For Label + series, this constraint is enforced separately for the category and series expressions.

For image interventions, the patched supports include the y-axis scale region, target-bar regions, two affected legend-entry regions, and two affected x-axis label slots. For prompt interventions, we patch the corresponding prompt-token positions.

For bar height, we change the target bar height by exactly one y-axis tick step and define three ROIs: Union, the union of the target-bar regions in the original and counterfactual images; Difference, the symmetric difference between the two target-bar regions, concentrated around the changed bar top; and Intersection, the intersection of the two target-bar regions, corresponding to the invariant shared bar body. These three conditions share identical counterfactual pairs and differ only in the patched ROI. Complete counterfactual construction and patched-support definitions for all nine analysis conditions are provided in Table~\ref{tab:interventions}.

\subsection{Evaluation Results on Original and Counterfactual Chart Value Reading}

Before analyzing internal activations, we evaluate the value-reading performance of both models on the original and counterfactual charts. Across 5,932 queries, Qwen2.5-VL-7B and InternVL3.5-8B achieve 97.57\% and 97.27\% accuracy on the original charts, respectively. For target-relevant bar-height, legend-swap, and x-label-swap counterfactuals, both endpoints are answered correctly in 95.03\%--95.52\% of the pairs. Axis-scale changes are more challenging, with both-endpoint correctness of 72.02\% for Qwen and 84.95\% for InternVL. In contrast, non-target edits preserve the original correctness status in 99.36\%--99.90\% of pairs.

These results indicate that the models generally maintain accurate value reading under local target-relevant changes, while edits unrelated to the queried value have minimal impact on correctness. Axis-scale changes present a substantially greater challenge. Full results are provided in Appendix~\ref{sec:chartvaluereadingdatagendetail}.

\section{Activation Patching Framework}

\subsection{Activation patching and metric}
We score complete answer token sequences with teacher forcing. This avoids reducing multi-token numbers to their first token. For an input $x$, query $q$, and answer sequence $a=(a_1,\ldots,a_T)$,
\begin{equation}
S_x(a)=\sum_{t=1}^{T}\log P(a_t\mid x,q,a_{<t}).
\end{equation}
Let $a_o$ and $a_c$ be the original and counterfactual gold answers. Define the original, counterfactual, and patched margins:
\begin{align}
M_o &= S_o(a_o)-S_o(a_c),\\
M_c &= S_c(a_o)-S_c(a_c),\\
M_p &= S_p(a_o)-S_p(a_c).
\end{align}
We report
\begin{equation}
R=\frac{M_p-M_c}{M_o-M_c}.
\label{eq:restore}
\end{equation}
$R=1$ restores the full endpoint margin gap and $R=0$ leaves the counterfactual preference unchanged. We do not clip $R$. No evaluated sample has $|M_o-M_c|<10^{-6}$.

All patches are original$\rightarrow$counterfactual. Image conditions patch the  projected image tokens, individual post-residual language layers, and contiguous windows of two or four layers. Prompt conditions patch changed prompt positions at individual layers and windows.

We use normalized restoration $R$ as the outcome measure for all
interventions (Eq.~\ref{eq:app_normalized_restoration}). For image
interventions, \emph{projected-site restoration} patches the
ROI-selected visual tokens at the projected image-token site
(Eq.~\ref{eq:app_projected_restoration}), whereas
\emph{layer-wise restoration} applies the same patch at individual
language layers (Eq.~\ref{eq:app_layer_restoration}). Prompt
interventions patch the changed category or series token positions at
each language layer and therefore have no projected-site image
condition (Eq.~\ref{eq:app_prompt_restoration}).

\subsection{Models and spatial mappings}

We evaluate \qwen-7B-Instruct
\citep{bai2025qwen25vltechnicalreport} and \intern-8B
\citep{wang2025internvl35advancingopensourcemultimodal}.
\qwen{} produces a $96{\times}96$ vision grid, which is merged into a
$48{\times}48$ image-token grid before its 28 language layers.
\intern{} similarly uses a $32{\times}32$ vision grid, a
$16{\times}16$ merged image-token grid, and 36 language layers. We do
not patch the fine vision grids; all image interventions operate on
merged image-token positions.

For each image ROI in Table~\ref{tab:interventions}, the
\textbf{any-overlap} rule selects every merged image token whose spatial
cell overlaps at least one ROI pixel. This gives a set of positions on
the $48{\times}48$ grid for \qwen{} and on the $16{\times}16$ grid for
\intern{}.

In some experiment setting of \qwen{}, we additionally evaluate a \textbf{$3{\times}3$ coarse}
mapping. Each token selected by any-overlap is expanded to its containing
non-overlapping $3{\times}3$ block on the merged grid. This includes
nearby context and reduces sensitivity to token boundaries; it is used
as a robustness condition rather than as an estimate of a minimal
causal region. \intern{} uses only any-overlap.

The selected merged-token positions are shared across patching sites.
For \emph{projected-site restoration}, they are patched once at the
projected image-token representation entering the language model. For
\emph{layer-wise restoration}, the same positions are patched separately
at each language layer. Thus, the two analyses differ in where the
states are replaced, not in which spatial tokens represent the ROI.
Prompt interventions do not use these spatial mappings and patch only
the designated prompt-token positions.

\section{Factor-Wise Casual Localization}
In this section we mainly analysis the single-factor defined in Table~\ref{tab:cf_behavior}

\subsection{Experiment Setup}

\paragraph{Evaluation Data Auditing}
\label{sec:eval_data_audit}

The 500 source charts yield 5,932 value-reading queries.
For each counterfactual condition, we retain only pairs where
both models correctly answer the original and counterfactual
inputs, and the two correct answers differ.
Thus, all selected examples are successfully solved by
both Qwen2.5-VL and InternVL3.5 before activation patching.

We select 500 examples for each of the nine conditions.
To make comparisons fair, related conditions use the same
original queries. Specifically, the three bar-height regions
(Union, Difference, and Intersection), legend swap, and
x-label swap share the same 500 original queries.
The three bar-height conditions also use exactly the same
counterfactual chart pairs, differing only in which region
is patched. Similarly, the three prompt conditions
(Label, Series, and Label + Series) share another set of
500 original queries. Axis scale is evaluated on a
separate set of 500 queries.
For each condition, both models are evaluated on the
same examples.
Further details on data
selection are provided in Appendix~\ref{app:eval_data_audit}.

\paragraph{Related Evaluation Statistic Metrics}
The normalized restoration R are used to compute the statistic metrics.
We report the query-weighted mean and sometimes a
5,000-resample bootstrap 95\% confidence interval
(Eqs.~\ref{eq:app_query_mean}--\ref{eq:app_bootstrap_ci}). Conditions
evaluated on the same queries are compared using within-query paired
differences (Eqs.~\ref{eq:app_paired_difference}--%
\ref{eq:app_paired_mean}). Full experiments and their related statistics metrics are in appendix~\ref{sec:app_singlefactorprojectsitechangebar}.

\subsection{Projected-site normalized restoration Analysis}
\subsubsection{Overall restoration across visual interventions}
Table~\ref{tab:allprojected} provides the complete projected-site overview. Whole-bar, legend, and InternVL x-label patches approach full restoration. Axis and x-label restoration vary more strongly with architecture and token mapping. And more discussions with coarse mapping results are in appendix~\ref{sec:app_singlefactorprojectsitechangebar}.

\subsubsection{Changed-Height Bar ROI Patching Produces Higher Restoration Than Invariant-Body Patching}

Table~\ref{tab:allprojected} reports projected-site restoration. The ordering is identical in all settings:
$R_{\unionroi}>R_{\diffroi}\gg R_{\introi}$.
Mean \diffroi{} restoration is $0.783$, $0.912$, and $0.936$ for \qwen{} any-overlap, \qwen{} coarse, and \intern{}, compared with $0.328$, $0.420$, and $0.490$ for \introi{}. The paired \diffroi--\introi{} gaps are $0.455$ [0.430, 0.478], $0.492$ [0.470, 0.513], and $0.445$ [0.420, 0.471]. The direction holds for 94.6\%, 97.6\%, and 95.6\% of queries, respectively, and for 89.6\% of queries simultaneously across all three settings. More discussions and robustness checks are in appendix~\ref{sec:app_singlefactorprojectsitechangebar}.

\begin{figure*}[t]
\centering
\includegraphics[width=0.98\textwidth]{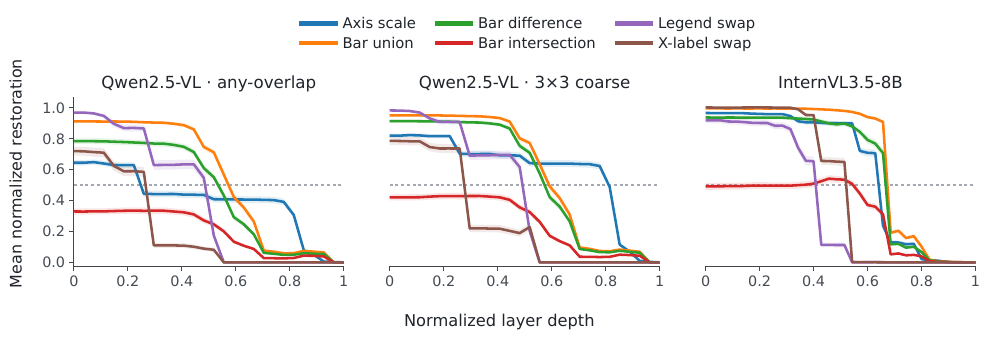}
\caption{Mean single-layer restoration for image-based counterfactual interventions. The absolute heights reflect both initial and later accessibility of each patch ROI.}
\label{fig:layers}
\end{figure*}

\begin{table}[t]
\centering
\footnotesize
\setlength{\tabcolsep}{3pt}
\renewcommand{\arraystretch}{0.95}
\begin{tabular}{@{}lcccccc@{}}
\toprule
Setting & Axis & Union & Diff. & Inter. & Leg. & X-lab. \\
\midrule
Qwen any         & .643 & .910 & .783 & .328 & .968 & .717 \\
Qwen $3{\times}3$ & .819 & .950 & .912 & .420 & .981 & .784 \\
InternVL         & .965 & .994 & .936 & .490 & .918 & 1.000 \\
\bottomrule
\end{tabular}
\caption{Mean normalized restoration at projected sites
for image interventions ($n=500$ each).
Union, Diff., and Inter. refer to bar operations.
Prompt interventions have no projected-image condition.}
\label{tab:allprojected}
\end{table}

\subsection{Layer-wise Analysis}
In section we present our main findings, more experiements are in the appendix~\ref{sec:app_singlefactorlayerwiseanalysis}.

\paragraph{Restoration from legend- and series-related positions declines at earlier layers than restoration from bar-height and axis-scale positions.}
Figure~\ref{fig:layers} compares single-layer restoration across normalized language-model depth. In both models, legend-swap restoration declines earlier than bar-height and axis-scale restoration, while the relative patterns x-label interventions vary more across architectures. More statistic analysis are in appendix~\ref{sec:app_singlefactorlayerwiseanalysis}.

\section{Causal role of prompt-series states.}
\label{sec:legend_handoff}

The single-factor analysis in last section shows that local restoration from visual-legend positions declines earlier across language-model depth than restoration from bar geometry and axis scale. This raises a further question:  When local restoration from visual-legend positions declines, can other states still restore answer preference under legend swaps, and do these states causally contribute to the downstream effect of an earlier visual-legend patch?  All statistical analyses in this section use normalized restoration, $R$, as the outcome measure.

\subsection{Data}
\label{sec:legend_handoff_data}

\paragraph{Evaluation data auditing.}
We reuse the 500 counterfactual queries from the \emph{Legend swap} condition constructed in the previous section. We split them by source chart into a pilot set of 100 queries and a held-out set of 400 queries. The two sets contain no overlapping source charts, preventing the same chart from contributing to both layer selection and the final effect estimates.

\subsection{Identifying Candidate States}
\label{sec:legend_destination_patching}

\subsubsection{Experimental Setup}

We first examine which token positions and layers contain states that can restore the original answer preference under a legend swap.

For each language-model layer, we use the counterfactual input as the receiver computation and replace the activation at a candidate location with the corresponding activation from the original-input forward pass. Following the terminology of Table~1, the candidate locations are the two affected visual legend-entry regions, including their visual entries and text labels, the target-bar region, the prompt-series token positions, and the final prompt token. The target-bar region is the union of the two bar regions at the queried category for the two series involved in the legend swap. The prompt-label token positions and a matched control-bar region (union of tow bars at other x-axis label position) serve as controls.

This analysis uses only the 400 held-out queries. The layer-wise curves use 500 query-level bootstrap resamples, and the shaded bands show pointwise 95\% confidence intervals.

\subsubsection{Results}

\begin{figure}[t]
    \centering
    \includegraphics[width=0.98\columnwidth]{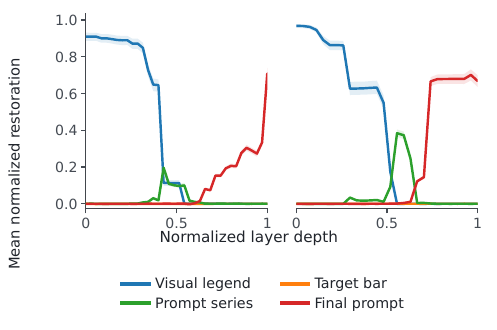}

    \caption{Layer-wise normalized restoration at the main candidate locations for InternVL3.5-8B (left) and Qwen2.5-VL-7B (right). Shaded bands denote pointwise 95\% confidence intervals from 500 query-level bootstrap resamples.}
    \label{fig:legend_destination_profiles}
\end{figure}

\paragraph{Middle-layer prompt-series states show restoration effects under legend swaps.}
Figure~\ref{fig:legend_destination_profiles} compares the layer-wise restoration profiles at the main candidate locations. Both models show a similar broad pattern: restoration at the visual-legend positions is strongest in early layers and declines with depth, while restoration at the prompt-series positions becomes substantial in middle layers. Restoration at the final prompt position is more prominent in later layers. In contrast, the target-bar region shows little restoration across layers. These results indicate that prompt-series states can restore the original answer preference at depths where restoration from the visual-legend positions has become weak. However, the different restoration profiles do not by themselves establish information transfer between these locations. We also examine the effect of the matched controls with the results presented in appendix~\ref{sec:casualroleindentifycandidates}.

\begin{table*}[t]
    \centering
    \small
    \resizebox{\textwidth}{!}{%
    \begin{tabular}{llcccc}
        \toprule
        & & \multicolumn{2}{c}{InternVL3.5-8B} & \multicolumn{2}{c}{Qwen2.5-VL-7B} \\
        \cmidrule(lr){3-4}\cmidrule(lr){5-6}
        Reset type & Reset location
        & Post-reset $R$ [95\% cluster CI] & Blocking effect $\Delta R$
        & Post-reset $R$ [95\% cluster CI] & Blocking effect $\Delta R$ \\
        \midrule
        No-reset baseline & None
        & 0.910 [0.886, 0.932] & --
        & 0.967 [0.954, 0.978] & -- \\
        Prompt reset & Prompt-series positions
        & 0.701 [0.676, 0.727] & \textbf{0.209}
        & 0.541 [0.518, 0.564] & \textbf{0.426} \\
        Prompt reset & Prompt-label positions
        & 0.909 [0.886, 0.932] & 0.001
        & 0.932 [0.918, 0.944] & 0.035 \\
        Visual reset & Target-bar region
        & 0.909 [0.886, 0.932] & 0.001
        & 0.967 [0.954, 0.978] & 0.000 \\
        Visual reset & Matched control-bar region
        & 0.910 [0.886, 0.932] & 0.000
        & 0.967 [0.954, 0.978] & 0.000 \\
        \bottomrule
    \end{tabular}%
    }
    \caption{Reset-based blocking results on the 400 held-out queries. Post-reset $R$ is normalized restoration after source rescue and the specified destination reset. The blocking effect is the decrease relative to the no-reset source-rescue baseline.}
    \label{tab:legend_reset_results}
\end{table*}

\subsection{Testing Causal Dependence}
\label{sec:legend_reset_mediation}

\subsubsection{Experimental Setup}

We test whether prompt-series states contribute to the downstream effect of an earlier visual-legend patch through a two-stage intervention. Starting from the counterfactual input, we first restore the original activations at the affected visual-legend positions (source rescue), then reset the prompt-series activations at a deeper layer to their counterfactual values. We measure the reduction in normalized restoration as the blocking effect, $\Delta R_d = R_{\mathrm{source\ only}} - R_{\mathrm{source+reset}(d)}$, where a positive value indicates attenuation of source rescue.

For each model, the source and reset layers are selected using 100 pilot queries by maximizing source rescue and its subsequent reduction, respectively. The selected layers are fixed before evaluation on 400 held-out queries, with no source-chart overlap between splits. We compare prompt-series resets against prompt-label, target-bar, and matched control-bar resets to assess location specificity. All effects are estimated on the held-out set, with 95\% confidence intervals obtained from 5,000 source-chart cluster-bootstrap resamples (Eq~\ref{eq:app_cluster_bootstrap}). Further implementation details are provided in Appendix~\ref{sec:testingcasualdependencydetails}.

\subsubsection{Results}
\textbf{Prompt-series states contribute to the downstream effect of an earlier visual-legend patch.}
As shown in Table~\ref{tab:legend_reset_results}, resetting the prompt-series states substantially reduces the source-rescue effect in both models. In contrast, resetting the prompt-label positions produces much smaller reductions, while resetting either the target-bar region or the matched control-bar region has negligible effects. These results show that the effect of the earlier visual-legend patch is selectively sensitive to the later prompt-series states among the tested locations. Together with the single-location restoration results, these findings suggest that prompt-series states play a causal role in the downstream effects of an earlier visual-legend intervention.

\section{Factorial Recombination of Bar Geometry and Axis Scale}
\label{sec:factorial-recombination}

\begin{table}[t]
\centering
\scriptsize
\begin{tabular}{llll}
\toprule
Chart & Bar geometry & Axis scale & Correct answer \\
\midrule
$C_{00}$ & Base height      & Base scale      & $v$ \\
$C_{10}$ & Alternate height & Base scale      & $v+s$ \\
$C_{01}$ & Base height      & Alternate scale & $2v$ \\
$C_{11}$ & Alternate height & Alternate scale & $2(v+s)$ \\
\bottomrule
\end{tabular}
\caption{The four charts in a factorial family. The first index denotes bar-geometry state and the second denotes axis-scale state; 0 is the base state and 1 is the alternate state.}
\label{tab:factorial-cells}
\end{table}

\begin{table*}[t]
\centering
\tiny
\resizebox{\textwidth}{!}{%
\begin{tabular}{llccc}
\toprule
Model & Intervention & Four-candidate target accuracy [95\% CI] & $R_{\mathrm{joint\text{-}cross}}$ [95\% CI] & $R_{\mathrm{joint\text{-}coherent}}$ [95\% CI] \\
\midrule
InternVL3.5-8B & Geometry-only  & 0.996 [0.990, 1.000] & -- & -- \\
InternVL3.5-8B & Scale-only     & 1.000 [1.000, 1.000] & -- & -- \\
InternVL3.5-8B & Joint-cross    & 0.996 [0.990, 1.000] & 0.977 [0.973, 0.980] & -- \\
InternVL3.5-8B & Joint-coherent & 0.996 [0.990, 1.000] & -- & 0.977 [0.974, 0.981] \\
Qwen2.5-VL-7B  & Geometry-only  & 0.974 [0.960, 0.986] & -- & -- \\
Qwen2.5-VL-7B  & Scale-only     & 0.994 [0.986, 1.000] & -- & -- \\
Qwen2.5-VL-7B  & Joint-cross    & 0.908 [0.882, 0.932] & 0.803 [0.795, 0.811] & -- \\
Qwen2.5-VL-7B  & Joint-coherent & 0.996 [0.990, 1.000] & -- & 0.867 [0.860, 0.873] \\
\bottomrule
\end{tabular}
}
\caption{Four-candidate target accuracy for all four interventions at the projected image-token site, together with the normalized restoration score for each joint intervention.}
\label{tab:factorial-projected-results}
\end{table*}

\subsection{Experimental Setup}
\label{sec:factorial-setup}

This section tests whether VLMs can combine bar-height and y-axis-scale internal states from different charts to recover the value implied by their combination.

Each family contains four charts formed by crossing two target-bar heights with two axis-scale settings. One chart is used as the receiver. Relative to the receiver, the \emph{geometry donor} changes only the target-bar height while keeping the axis scale fixed, and the \emph{scale donor} changes only the axis scale while preserving the rendered target-bar height. The remaining chart, called the \emph{diagonal}, contains both alternative factors and therefore represents the target combination of bar geometry and axis scale.

We apply four interventions during the receiver forward pass while keeping the prompt unchanged:
(1) \textbf{Geometry-only} replaces activations at the target-bar region with those from the geometry donor. This condition tests whether the transferred bar-geometry information can affect the answer on its own.
(2) \textbf{Scale-only} replaces activations at the y-axis scale region with those from the scale donor. This condition tests whether the transferred axis-scale information can affect the answer on its own.
(3) \textbf{Joint-cross} takes target-bar-region activations from the geometry donor and y-axis-scale-region activations from the scale donor, and inserts both into the same receiver forward pass. It tests whether information from two different donor forward passes can be combined to recover a target answer that neither donor provides by itself.
(4) \textbf{Joint-coherent} takes both sets of activations from the diagonal and inserts them together into the same receiver forward pass. Comparing this condition with joint-cross tests whether joint restoration depends on the two activation sets coming from the same image context.

We perform these interventions both at the projected image-token site and at individual language layers.

\subsection{Evaluation Metrics}
\label{sec:factorial-metrics}

We score all four candidate answers using complete-sequence log probabilities. Four-candidate target accuracy is the fraction of families in which the designated target strictly out-scores the other candidates: the geometry-donor answer for geometry-only, the scale-donor answer for scale-only, and the diagonal answer for both joint interventions. For the joint interventions, we additionally measure normalized restoration:

$$
R_j=\frac{M_{\mathrm{patch}}^j-M_{\mathrm{cf}}}{M_{\mathrm{ori}}-M_{\mathrm{cf}}}
$$

where the margins compare the diagonal and receiver answers. Bootstrap 95\% confidence intervals use 5,000 family-level resamples for projected-site results and 500 for layer-wise results. Further details are provided in Appendix~\ref{sec:fraemd}.

\subsection{Data Generation and Auditing}
\label{sec:factorial-data}

We generate 10,000 factorial chart families by crossing two target-bar heights (one tick apart) with two y-axis scales (differing by a factor of two), yielding four distinct answers (Table~\ref{tab:factorial-cells}). Geometry and scale ROIs are non-overlapping. We retain 4,903 families for which both models correctly answer all four unpatched charts, and select 500 balanced by receiver type and base tick interval, without using patching outcomes. Our conclusions therefore apply to successful unpatched value-reading cases. Generation, filtering, and data auditing details are provided in Appendix~\ref{sec:fredaa}.

\subsection{Projected-Site Composition Analysis}
\label{sec:factorial-projected}

We first apply all four interventions at the projected image-token site. Four-candidate target accuracy indicates whether the predefined target answer has the highest sequence score. For the two joint interventions, $R_{\mathrm{joint\text{-}cross}}$ and $R_{\mathrm{joint\text{-}coherent}}$ measure how much the intervention moves the answer preference from the receiver (cf) baseline toward the diagonal (ori) baseline. The results are shown in Table~\ref{tab:factorial-projected-results}.

\textbf{InternVL3.5-8B combines bar-geometry and axis-scale information equally well whether the two activation sets come from separate donors or from the same diagonal input.} Geometry-only and scale-only both achieve high target accuracy, showing that either activation set can independently shift the model toward the corresponding single-factor donor answer. Joint-cross and joint-coherent have the same target-accuracy point estimate and nearly identical normalized restoration, indicating little sensitivity to whether the two activation sets share the same source context.

\textbf{Qwen2.5-VL-7B can also combine bar-geometry and axis-scale information across forward passes, but it is more sensitive to whether the two activation sets come from the same input.} Geometry-only and scale-only again achieve high target accuracy. However, joint-cross has lower target accuracy and normalized restoration than joint-coherent. Thus, Qwen selects the diagonal target in most joint-cross families, but combines the two activation sets less effectively when they come from separate geometry and scale donors than when both come from the diagonal.

\subsection{Layer-Wise Composition Analysis}
\label{sec:factorial-layerwise}

Joint restoration decreases with language-model depth in both models (Fig~\ref{fig:factorial-layerwise-restoration}). Joint-cross and joint-coherent exhibit similar depth profiles in InternVL3.5-8B, whereas Qwen2.5-VL-7B shows lower joint-cross restoration and an earlier decline. This pattern is consistent with Qwen's greater sensitivity to donor context.

\section{Conclusion}

We use counterfactual activation patching to study exact bar-value
reading in two VLMs. Single-factor interventions show stronger
restoration from changed bar tops than unchanged bodies, while
legend- and series-related states lose local restoration earlier
than geometry and scale states. Under legend swaps, middle-layer
prompt-series states restore answer preference, and resetting them
selectively weakens restoration from an earlier visual-legend patch.
Both models combine geometry and scale states from separate donors,
but Qwen is more sensitive to donor context than InternVL.
Together, these findings provide causal evidence about the internal
states supporting exact bar-value reading.

\section*{Limitations}
The study is conditional on strict four-endpoint success: both models must answer original and counterfactual inputs correctly. This yields a clean margin-based intervention metric but does not characterize failure cases. The charts are synthetic, use fixed layouts, and cover only vertical bars with integer values; natural charts introduce OCR noise, irregular styling, truncated axes, annotations, and broader semantic priors.

\bibliographystyle{acl_natbib}
\bibliography{custom}

@inproceedings{masry-etal-2022-chartqa,
    title = "{C}hart{QA}: A Benchmark for Question Answering about Charts with Visual and Logical Reasoning",
    author = "Masry, Ahmed  and
      Long, Do Xuan  and
      Tan, Jia Qing  and
      Joty, Shafiq  and
      Hoque, Enamul",
    editor = "Muresan, Smaranda  and
      Nakov, Preslav  and
      Villavicencio, Aline",
    booktitle = "Findings of the Association for Computational Linguistics: ACL 2022",
    month = may,
    year = "2022",
    address = "Dublin, Ireland",
    publisher = "Association for Computational Linguistics",
    url = "https://aclanthology.org/2022.findings-acl.177/",
    doi = "10.18653/v1/2022.findings-acl.177",
    pages = "2263--2279"
}

@misc{kahou2018figureqaannotatedfiguredataset,
      title={FigureQA: An Annotated Figure Dataset for Visual Reasoning}, 
      author={Samira Ebrahimi Kahou and Vincent Michalski and Adam Atkinson and Akos Kadar and Adam Trischler and Yoshua Bengio},
      year={2018},
      eprint={1710.07300},
      archivePrefix={arXiv},
      primaryClass={cs.CV},
      url={https://arxiv.org/abs/1710.07300}, 
}

@misc{methani2020plotqareasoningscientificplots,
      title={PlotQA: Reasoning over Scientific Plots}, 
      author={Nitesh Methani and Pritha Ganguly and Mitesh M. Khapra and Pratyush Kumar},
      year={2020},
      eprint={1909.00997},
      archivePrefix={arXiv},
      primaryClass={cs.CV},
      url={https://arxiv.org/abs/1909.00997}, 
}

@misc{masry2025chartqaprodiversechallengingbenchmark,
      title={ChartQAPro: A More Diverse and Challenging Benchmark for Chart Question Answering}, 
      author={Ahmed Masry and Mohammed Saidul Islam and Mahir Ahmed and Aayush Bajaj and Firoz Kabir and Aaryaman Kartha and Md Tahmid Rahman Laskar and Mizanur Rahman and Shadikur Rahman and Mehrad Shahmohammadi and Megh Thakkar and Md Rizwan Parvez and Enamul Hoque and Shafiq Joty},
      year={2025},
      eprint={2504.05506},
      archivePrefix={arXiv},
      primaryClass={cs.CL},
      url={https://arxiv.org/abs/2504.05506}, 
}

@misc{wang2024charxivchartinggapsrealistic,
      title={CharXiv: Charting Gaps in Realistic Chart Understanding in Multimodal LLMs}, 
      author={Zirui Wang and Mengzhou Xia and Luxi He and Howard Chen and Yitao Liu and Richard Zhu and Kaiqu Liang and Xindi Wu and Haotian Liu and Sadhika Malladi and Alexis Chevalier and Sanjeev Arora and Danqi Chen},
      year={2024},
      eprint={2406.18521},
      archivePrefix={arXiv},
      primaryClass={cs.CL},
      url={https://arxiv.org/abs/2406.18521}, 
}

@misc{xu2024chartbenchbenchmarkcomplexvisual,
      title={ChartBench: A Benchmark for Complex Visual Reasoning in Charts}, 
      author={Zhengzhuo Xu and Sinan Du and Yiyan Qi and Chengjin Xu and Chun Yuan and Jian Guo},
      year={2024},
      eprint={2312.15915},
      archivePrefix={arXiv},
      primaryClass={cs.CV},
      url={https://arxiv.org/abs/2312.15915}, 
}

@article{Mukherjee_2026,
   title={EncQA: Benchmarking Vision-Language Models on Visual Encodings for Charts},
   volume={32},
   ISSN={2160-9306},
   url={http://dx.doi.org/10.1109/TVCG.2025.3634249},
   DOI={10.1109/tvcg.2025.3634249},
   number={1},
   journal={IEEE Transactions on Visualization and Computer Graphics},
   publisher={Institute of Electrical and Electronics Engineers (IEEE)},
   author={Mukherjee, Kushin and Ren, Donghao and Moritz, Dominik and Assogba, Yannick},
   year={2026},
   month=Jan, pages={648–658} }

@misc{tartaglini2025diagnosingbottlenecksdatavisualization,
      title={Diagnosing Bottlenecks in Data Visualization Understanding by Vision-Language Models}, 
      author={Alexa R. Tartaglini and Satchel Grant and Daniel Wurgaft and Christopher Potts and Judith E. Fan},
      year={2025},
      eprint={2510.21740},
      archivePrefix={arXiv},
      primaryClass={cs.CV},
      url={https://arxiv.org/abs/2510.21740}, 
}

@misc{meng2023locatingeditingfactualassociations,
      title={Locating and Editing Factual Associations in GPT}, 
      author={Kevin Meng and David Bau and Alex Andonian and Yonatan Belinkov},
      year={2023},
      eprint={2202.05262},
      archivePrefix={arXiv},
      primaryClass={cs.CL},
      url={https://arxiv.org/abs/2202.05262}, 
}

@inproceedings{NEURIPS2023_34e1dbe9,
 author = {Conmy, Arthur and Mavor-Parker, Augustine and Lynch, Aengus and Heimersheim, Stefan and Garriga-Alonso, Adri\`{a}},
 booktitle = {Advances in Neural Information Processing Systems},
 doi = {10.52202/075280-0719},
 editor = {A. Oh and T. Naumann and A. Globerson and K. Saenko and M. Hardt and S. Levine},
 pages = {16318--16352},
 publisher = {Curran Associates, Inc.},
 title = {Towards Automated Circuit Discovery for Mechanistic Interpretability},
 url = {https://proceedings.neurips.cc/paper_files/paper/2023/file/34e1dbe95d34d7ebaf99b9bcaeb5b2be-Paper-Conference.pdf},
 volume = {36},
 year = {2023}
}

@misc{zhang2024bestpracticesactivationpatching,
      title={Towards Best Practices of Activation Patching in Language Models: Metrics and Methods}, 
      author={Fred Zhang and Neel Nanda},
      year={2024},
      eprint={2309.16042},
      archivePrefix={arXiv},
      primaryClass={cs.LG},
      url={https://arxiv.org/abs/2309.16042}, 
}

@misc{salazar2026pathwaysvisualinformationflow,
      title={Pathways of Visual Information Flow in Vision-Language Models}, 
      author={Israfel Salazar and Stella Frank and Dan Oneata and Desmond Elliott and Constanza Fierro},
      year={2026},
      eprint={2607.03358},
      archivePrefix={arXiv},
      primaryClass={cs.CV},
      url={https://arxiv.org/abs/2607.03358}, 
}

@misc{palit2023visionlanguagemechanisticinterpretabilitycausal,
      title={Towards Vision-Language Mechanistic Interpretability: A Causal Tracing Tool for BLIP}, 
      author={Vedant Palit and Rohan Pandey and Aryaman Arora and Paul Pu Liang},
      year={2023},
      eprint={2308.14179},
      archivePrefix={arXiv},
      primaryClass={cs.CL},
      url={https://arxiv.org/abs/2308.14179}, 
}

@misc{golovanevsky2025vlmsnoticemechanisticinterpretability,
      title={What Do VLMs NOTICE? A Mechanistic Interpretability Pipeline for Gaussian-Noise-free Text-Image Corruption and Evaluation}, 
      author={Michal Golovanevsky and William Rudman and Vedant Palit and Ritambhara Singh and Carsten Eickhoff},
      year={2025},
      eprint={2406.16320},
      archivePrefix={arXiv},
      primaryClass={cs.CL},
      url={https://arxiv.org/abs/2406.16320}, 
}

@misc{wang2025internvl35advancingopensourcemultimodal,
      title={InternVL3.5: Advancing Open-Source Multimodal Models in Versatility, Reasoning, and Efficiency}, 
      author={Weiyun Wang and Zhangwei Gao and Lixin Gu and Hengjun Pu and Long Cui and Xingguang Wei and Zhaoyang Liu and Linglin Jing and Shenglong Ye and Jie Shao and Zhaokai Wang and Zhe Chen and Hongjie Zhang and Ganlin Yang and Haomin Wang and Qi Wei and Jinhui Yin and Wenhao Li and Erfei Cui and Guanzhou Chen and Zichen Ding and Changyao Tian and Zhenyu Wu and Jingjing Xie and Zehao Li and Bowen Yang and Yuchen Duan and Xuehui Wang and Zhi Hou and Haoran Hao and Tianyi Zhang and Songze Li and Xiangyu Zhao and Haodong Duan and Nianchen Deng and Bin Fu and Yinan He and Yi Wang and Conghui He and Botian Shi and Junjun He and Yingtong Xiong and Han Lv and Lijun Wu and Wenqi Shao and Kaipeng Zhang and Huipeng Deng and Biqing Qi and Jiaye Ge and Qipeng Guo and Wenwei Zhang and Songyang Zhang and Maosong Cao and Junyao Lin and Kexian Tang and Jianfei Gao and Haian Huang and Yuzhe Gu and Chengqi Lyu and Huanze Tang and Rui Wang and Haijun Lv and Wanli Ouyang and Limin Wang and Min Dou and Xizhou Zhu and Tong Lu and Dahua Lin and Jifeng Dai and Weijie Su and Bowen Zhou and Kai Chen and Yu Qiao and Wenhai Wang and Gen Luo},
      year={2025},
      eprint={2508.18265},
      archivePrefix={arXiv},
      primaryClass={cs.CV},
      url={https://arxiv.org/abs/2508.18265}, 
}

@misc{bai2025qwen25vltechnicalreport,
      title={Qwen2.5-VL Technical Report}, 
      author={Shuai Bai and Keqin Chen and Xuejing Liu and Jialin Wang and Wenbin Ge and Sibo Song and Kai Dang and Peng Wang and Shijie Wang and Jun Tang and Humen Zhong and Yuanzhi Zhu and Mingkun Yang and Zhaohai Li and Jianqiang Wan and Pengfei Wang and Wei Ding and Zheren Fu and Yiheng Xu and Jiabo Ye and Xi Zhang and Tianbao Xie and Zesen Cheng and Hang Zhang and Zhibo Yang and Haiyang Xu and Junyang Lin},
      year={2025},
      eprint={2502.13923},
      archivePrefix={arXiv},
      primaryClass={cs.CV},
      url={https://arxiv.org/abs/2502.13923}, 
}

@misc{vteam2026glm5vturbonativefoundationmodel,
      title={GLM-5V-Turbo: Toward a Native Foundation Model for Multimodal Agents}, 
      author={V Team and Wenyi Hong and Xiaotao Gu and Ziyang Pan and Zhen Yang and Yuting Wang and Yue Wang and Yuanchang Yue and Yu Wang and Yanling Wang and Yan Wang and Xijun Liu and Wenmeng Yu and Weihan Wang and Wei Li and Shuaiqi Duan and Sheng Yang and Ruiliang Lv and Mingdao Liu and Lihang Pan and Ke Ning and Junhui Ji and Jinjiang Wang and Jing Chen and Jiazheng Xu and Jiale Zhu and Jiale Cheng and Ji Qi and Guobing Gan and Guo Wang and Cong Yao and Zijun Dou and Zihao Zhou and Zihan Wang and Zhiqi Ge and Zhijie Li and Zhenyu Hou and Zhao Xue and Zehui Wang and Zehan Qi and Zehai He and Yutao Zhang and Yusen Liu and Yukuo Cen and Yuchen Li and Yuan Wang and Yu Yang and Yongbin Liu and Yijian Lu and Yifan Xu and Yanzi Wang and Yanxiao Zhao and Yanfeng Wang and Yadong Xue and Yabo Xu and Xinyu Zhang and Xinyu Liu and Xiao Liu and Wenyi Zhao and Wenkai Li and Tianyu Tong and Tianshu Zhang and Shudan Zhang and Shengdong Yan and Qinkai Zheng and Mingde Xu and Licheng Bao and lat Long long and Jiaxing Xu and Jiaxin Fan and Jiawen Qian and Jiali Chen and Jiahui Lin and Jiadai Sun and Haozhi Zheng and Haoran Wang and Haochen Li and Hanyu Lai and Han Xu and Fan Yang and Dan Zhang and Da Yin and Chuangxin Zhao and Chengcheng Wu and Boyan Shi and Bowen Lv and Bowei Jia and Bo Li and Bin Chen and Baoxu Wang and Peng Zhang and Debing Liu and Bin Xu and Juanzi Li and Minlie Huang and Yuxiao Dong and Jie Tang},
      year={2026},
      eprint={2604.26752},
      archivePrefix={arXiv},
      primaryClass={cs.CV},
      url={https://arxiv.org/abs/2604.26752}, 
}

@misc{liu2023deplotoneshotvisuallanguage,
      title={DePlot: One-shot visual language reasoning by plot-to-table translation}, 
      author={Fangyu Liu and Julian Martin Eisenschlos and Francesco Piccinno and Syrine Krichene and Chenxi Pang and Kenton Lee and Mandar Joshi and Wenhu Chen and Nigel Collier and Yasemin Altun},
      year={2023},
      eprint={2212.10505},
      archivePrefix={arXiv},
      primaryClass={cs.CL},
      url={https://arxiv.org/abs/2212.10505}, 
}

@misc{masry2023unichartuniversalvisionlanguagepretrained,
      title={UniChart: A Universal Vision-language Pretrained Model for Chart Comprehension and Reasoning}, 
      author={Ahmed Masry and Parsa Kavehzadeh and Xuan Long Do and Enamul Hoque and Shafiq Joty},
      year={2023},
      eprint={2305.14761},
      archivePrefix={arXiv},
      primaryClass={cs.CL},
      url={https://arxiv.org/abs/2305.14761}, 
}

@misc{liu2023matchaenhancingvisuallanguage,
      title={MatCha: Enhancing Visual Language Pretraining with Math Reasoning and Chart Derendering}, 
      author={Fangyu Liu and Francesco Piccinno and Syrine Krichene and Chenxi Pang and Kenton Lee and Mandar Joshi and Yasemin Altun and Nigel Collier and Julian Martin Eisenschlos},
      year={2023},
      eprint={2212.09662},
      archivePrefix={arXiv},
      primaryClass={cs.CL},
      url={https://arxiv.org/abs/2212.09662}, 
}

\appendix
\section{Appendix}

\subsection{Single-factor analysis More experiments and Details}

\paragraph{Full statistics evaluation metrics}
The normalized restoration R are used to compute the statistic metrics.
We report the query-weighted mean and sometimes a
5,000-resample bootstrap 95\% confidence interval
(Eqs.~\ref{eq:app_query_mean}--\ref{eq:app_bootstrap_ci}). Conditions
evaluated on the same queries are compared using within-query paired
differences (Eqs.~\ref{eq:app_paired_difference}--%
\ref{eq:app_paired_mean}).

We additionally use source-chart cluster
bootstrap and chart-equal estimates to account for multiple queries
derived from the same chart
(Eqs.~\ref{eq:app_cluster_bootstrap} and
\ref{eq:app_chart_equal}). The main bar-region comparison is also
repeated on queries for which the changed-top ROI selects no more
tokens than the invariant-body ROI
(Eqs.~\ref{eq:app_token_subset}--\ref{eq:app_token_control}).

For layer-wise analyses, layer depth is normalized as
$d_{\ell}=\ell/(L-1)$ (Eq.~\ref{eq:app_normalized_depth}). Raw AUC
summarizes total restoration across depth, while retention AUC
summarizes persistence after normalizing each profile by its layer-0
value (Eqs.~\ref{eq:app_raw_auc} and
\ref{eq:app_retention_auc}). Relative half-life is the first normalized
native-layer depth at which the mean retention profile reaches $0.5$
(Eqs.~\ref{eq:app_half_life_layer}--%
\ref{eq:app_relative_half_life}). Two- and four-layer windows test
whether restoration is distributed across neighboring layers
(Eq.~\ref{eq:app_window_restoration}). Spatial-mapping and cross-model
profile correlations are defined in
Sections~\ref{app:spatial_mapping_agreement} and
\ref{app:cross_model_corr}. Full metric definitions and computation
details are provided in Appendix~\ref{app:metric_definitions}.

\begin{figure}[t]
\centering
\includegraphics[width=0.8\columnwidth]{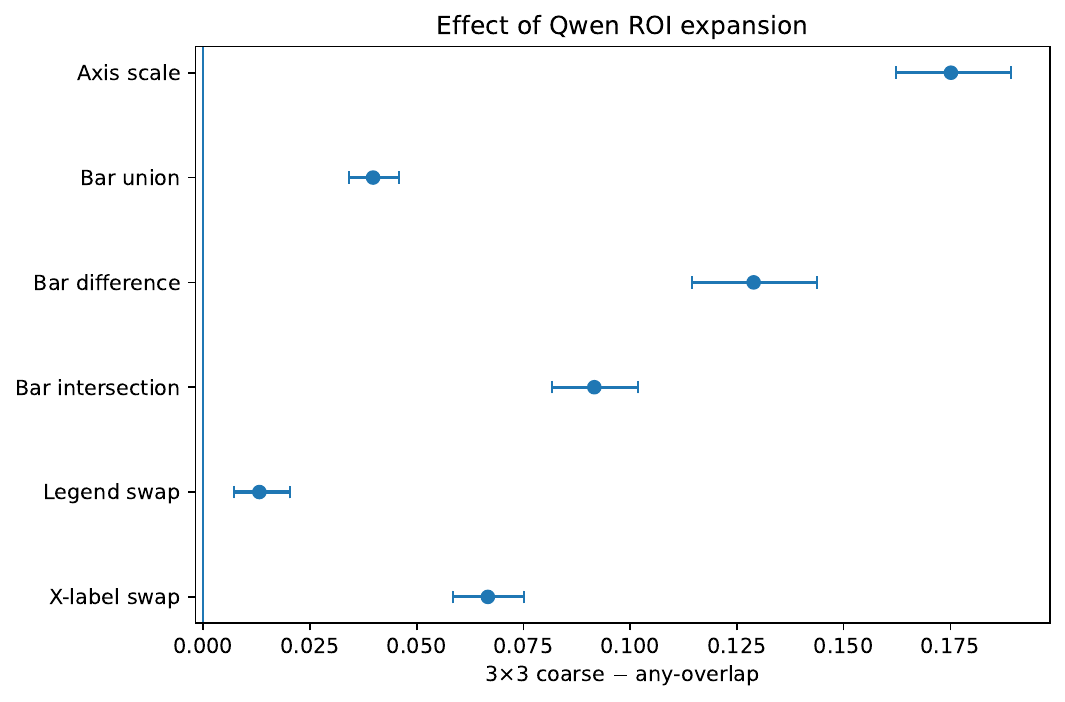}
\caption{Paired projected-restoration gain from \qwen{} $3{\times}3$ coarse neighborhoods over any-overlap.}
\label{fig:coarse}
\end{figure}

\subsubsection{Projected-site normalized restoration Analysis}

\paragraph{Spatial expansion changes magnitude}
The \qwen{} coarse mapping increases projected restoration for all image interventions (Figure~\ref{fig:coarse}). Gains are largest for axis scale (+0.175), the changed bar top (+0.129), the invariant body (+0.092), and x-labels (+0.067); legend gains only +0.013 because any-overlap is already near saturation. The restoration gain does not scale uniformly with the increase in selected token count across intervention types: legend tokens roughly double with little gain, whereas x-label tokens expand 2.61$\times$ but gain less than axis scale.

\paragraph{Changed-Height ROI Patching Produces Higher Restoration Than Invariant-Body Patching}
\label{sec:app_singlefactorprojectsitechangebar}
The Intersection ROI also produces nonzero restoration, despite excluding the changed-height band. However, its restoration effect is substantially smaller than that of the changed-height ROI under the tested counterfactuals.

The result is not explained by patching a larger region. Mean merged-token counts for \diffroi{} versus \introi{} are 21.8 vs. 53.2 (\qwen{} any), 47.0 vs. 97.6 (\qwen{} coarse), and 5.3 vs. 11.1 (\intern{}). Restricting to examples where \diffroi{} selects no more tokens than \introi{} yields gaps of $0.449$, $0.480$, and $0.438$. The \diffroi{} also captures most of the whole-bar effect: its mean sample-level \diffroi/\unionroi{} ratio is 0.857, 0.960, and 0.941; median ratios are 0.944, 0.994, and 0.993.

Source-chart dependence does not account for the effect. Cluster-bootstrap intervals for \diffroi--\introi{} are [0.431, 0.478], [0.471, 0.513], and [0.420, 0.471], nearly identical to query-level intervals. Chart-equal effects are 0.455, 0.490, and 0.441. Thus the conclusion holds whether the estimand weights queries or charts equally.

\subsubsection{Layer-wise analysis}
\label{sec:app_singlefactorlayerwiseanalysis}

\begin{figure*}[t]
\centering
\includegraphics[width=0.97\textwidth]{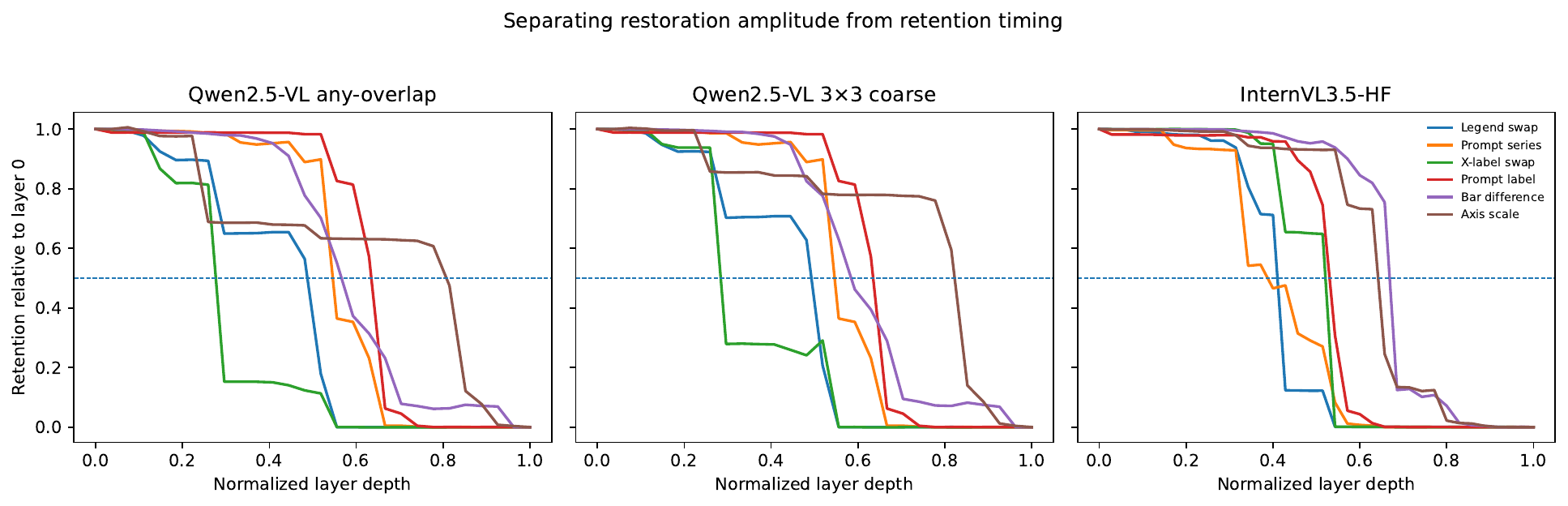}
\caption{Layer profiles after normalization by each intervention's layer-0 restoration. This view separates relative retention from the amount of information initially captured by an ROI.}
\label{fig:retention}
\end{figure*}

\paragraph{Restoration from legend- and series-related positions declines at earlier layers than restoration from bar-height and axis-scale positions.}
Absolute threshold crossings can be misleading because a strict ROI may start with a lower restoration ceiling. We therefore pair raw AUC with retention AUC. Relative to legend, bar difference has a raw-AUC advantage of 0.058 [0.042, 0.073] for \qwen{} any, 0.130 [0.118, 0.143] for \qwen{} coarse, and 0.251 [0.238, 0.264] for \intern{}. The corresponding retention-AUC advantages are 0.169, 0.174, and 0.259, with all intervals excluding zero. The bar-difference intervention has a larger raw restoration AUC and a higher retention AUC than the legend-swap intervention. These results indicate stronger restoration across the evaluated layers and a slower relative decline in its normalized restoration profile. Figure~\ref{fig:retention} visualizes this amplitude-normalized comparison.

Axis scale remains especially late in \qwen{}: its relative half-life occurs at normalized depth 0.815 (any) and 0.852 (coarse), versus approximately 0.59--0.63 for bar ROIs. In \intern{}, axis and bar half-lives are closer (0.657 and 0.686). We therefore do not claim a universal strict sequence of measuring height and then applying scale. The main observation is that restoration from legend- and series-related positions declines at earlier relative depths than restoration from bar-height and axis-scale positions.

\paragraph{Spatial expansion increases projected-site restoration magnitude while largely preserving the shapes of the layer-wise restoration profiles.}
Although the coarse mapping increases projected-site restoration
magnitude, it has little effect on the shape of the subsequent
language-layer restoration profiles. For each intervention, we average
the layer-wise restoration $R_{i,k,\ell}^{m}$ over queries separately
under the any-overlap and $3{\times}3$ coarse mappings, and compute the
within-model profile correlation $r_k^{\mathrm{map}}$ across Qwen's
native language layers
(Appendix~\ref{app:spatial_mapping_agreement};
Eqs.~\ref{eq:app_mapping_mean_profile}--%
\ref{eq:app_mapping_profile_corr}).
The mean correlation across all nine interventions is 0.995, with
correlations of 0.998 for bar difference, 0.9998 for bar intersection,
and 0.974 for axis scale. The prompt profiles are identical because
prompt-token positions do not depend on the visual ROI mapping.

\paragraph{The three bar ROIs show similar relative restoration profiles across depth despite their different restoration magnitudes. } This similarity describes the observed layer-wise effects but does not establish shared downstream computations.

\begin{figure*}[t]
\centering
\includegraphics[width=0.98\textwidth]{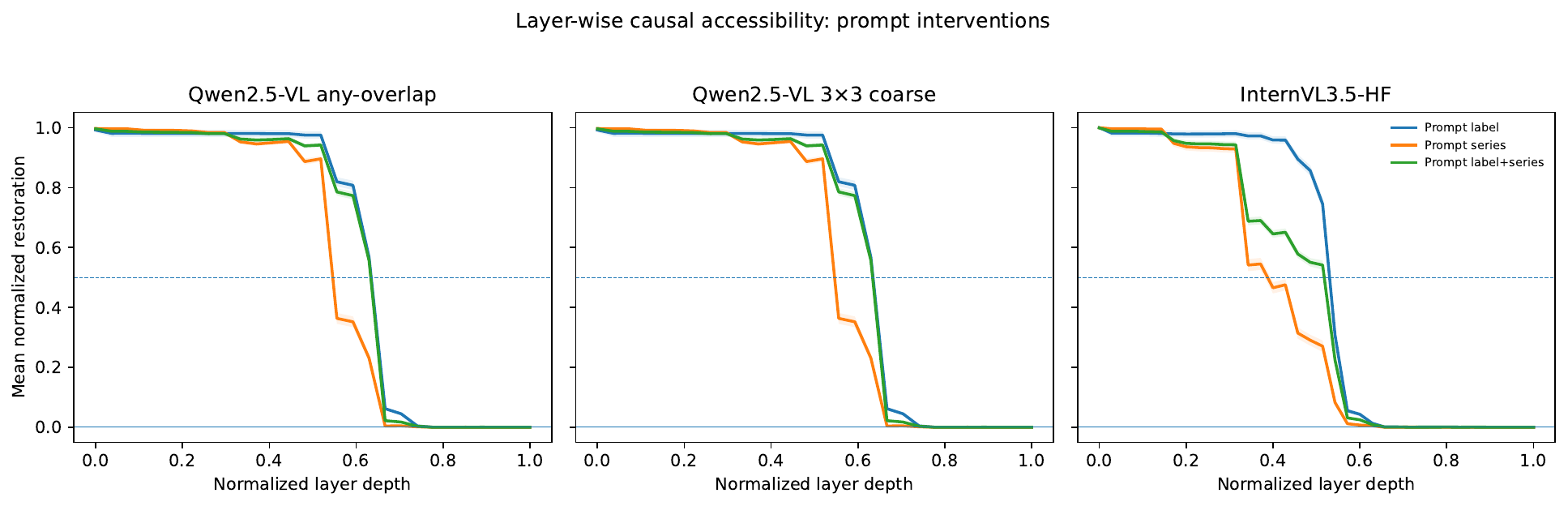}
\caption{Mean single-layer restoration for prompt-level representative interventions.}
\label{fig:layers_prompt}
\end{figure*}

\paragraph{The prompt interventions show that prompt-label and prompt-series interventions exhibit different layer-wise restoration profiles.} The specific results are shown in Figure~\ref{fig:layers_prompt}. Prompt-label interventions have a higher raw layer-profile
AUC than prompt-series interventions, with paired mean
advantages of 0.058 [0.049, 0.066] in \qwen{} and
0.113 [0.105, 0.121] in \intern{}. The paired difference
is positive for 96.2\% and 97.2\% of queries, respectively. Simultaneously replacing label and series yields an intermediate, non-additive profile. The combined prompt intervention does not exhibit a simple additive restoration profile relative to the individual prompt-label and prompt-series interventions.

\paragraph{The two models show similar average layer-wise restoration dynamics,
but substantially weaker agreement in projected-site restoration for
individual queries.}
 We first compare their language-layer profiles.
For each intervention $k$, we compute layer-wise image-token
restoration $R_{i,k,\ell}^{m}$ at every language layer and average it
over the 500 synchronized queries to obtain the native mean profile
$\overline{R}_{k,\ell}^{(m)}$
(Appendix~\ref{app:cross_model_corr};
Eqs.~\ref{eq:app_layer_restoration}
and~\ref{eq:app_native_mean_profile}).
Because \qwen{} and \intern{} have different numbers of language
layers, we map their native layers to normalized depth, linearly
interpolate the two mean profiles onto the same 101-point grid, and
compute the profile correlation
$r_k^{\mathrm{profile}}$
(Eqs.~\ref{eq:app_model_normalized_depth}--%
\ref{eq:app_profile_correlation}).
The resulting correlations are 0.951 for bar difference, 0.945 for
bar intersection, 0.934 for legend, and 0.799 for axis scale, with a
mean of approximately 0.898 across all nine interventions. Thus, after
averaging over queries, the two models tend to show similar changes in
local recoverability across relative language-layer depth.

We separately compare per-query projected-site restoration
$R_{i,k}^{\mathrm{proj},m}$, obtained by patching the ROI-selected
visual tokens at the projected image-token site
(Eq.~\ref{eq:app_projected_restoration}). This comparison does not use
language-layer profiles or depth interpolation: each query contributes
one projected-site restoration value from each model, and the two
length-500 vectors are correlated using
$r_k^{\mathrm{query}}$
(Eq.~\ref{eq:app_query_correlation}).
These correlations are much smaller: 0.365 for bar difference, 0.203
for bar intersection, and 0.168 for axis scale. The models therefore
share a similar aggregate pattern over language-layer depth, while
often differing in which individual chart queries are strongly
recoverable at the projected image-token site.

\begin{figure*}[p]
\centering
\includegraphics[width=0.98\textwidth]{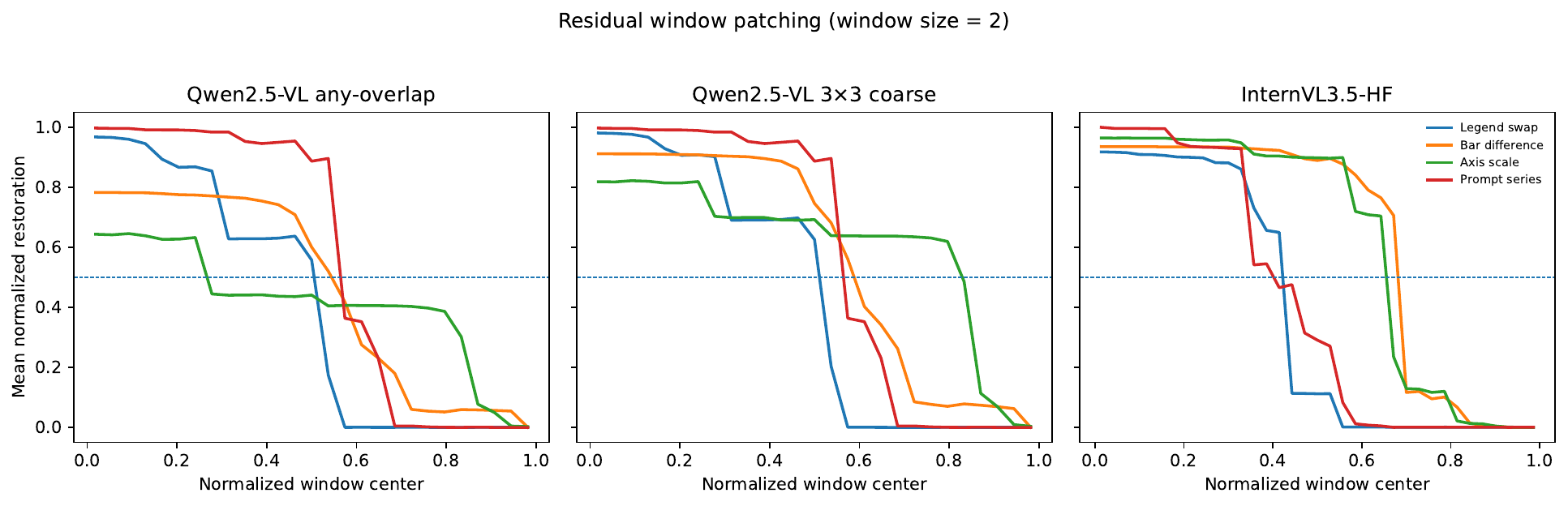}
\caption{Two-layer residual-window patching profiles.}
\label{fig:window2}
\end{figure*}

\begin{figure*}[p]
\centering
\includegraphics[width=0.98\textwidth]{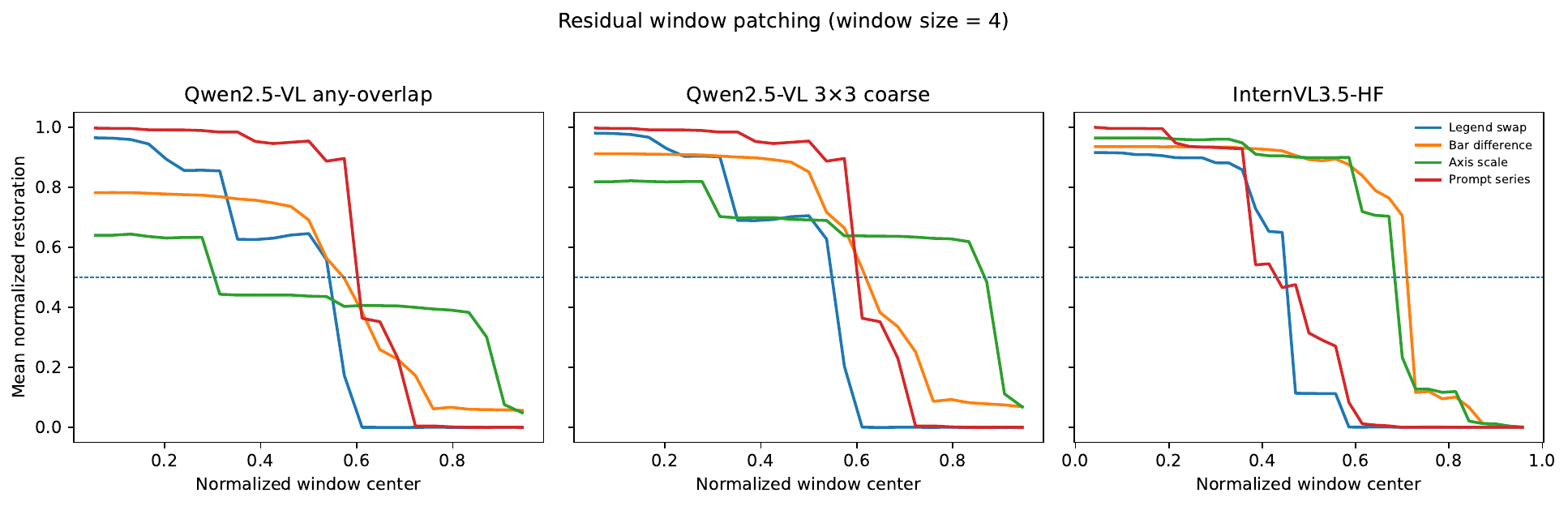}
\caption{Four-layer residual-window patching profiles.}
\label{fig:window4}
\end{figure*}

\paragraph{How Does Window Patching Affect Results}
A computation may be distributed across several neighboring
layers, so patching only one layer at a time could underestimate
how long the relevant information remains accessible. We
therefore repeat the analysis by jointly patching contiguous
windows of two or four layers. As expected, window patching
generally produces higher absolute restoration because more
hidden states are replaced. Importantly, however, it preserves
the main ordering and transition regions observed in the
single-layer analysis: window patching preserves the main depth-dependent restoration patterns observed in the single-layer analysis: legend-swap and prompt-series interventions show stronger restoration in earlier windows, while bar-height and axis-scale interventions retain substantial restoration at later windows.
Results are shown in Figure~\ref{fig:window2} and Figure~\ref{fig:window4}.

\subsection{Causal role of prompt-series states More experiments and Details}

\subsubsection{Identify Candidates Results}
\label{sec:casualroleindentifycandidates}
\begin{figure*}[t]
    \centering
    \includegraphics[width=0.49\textwidth]{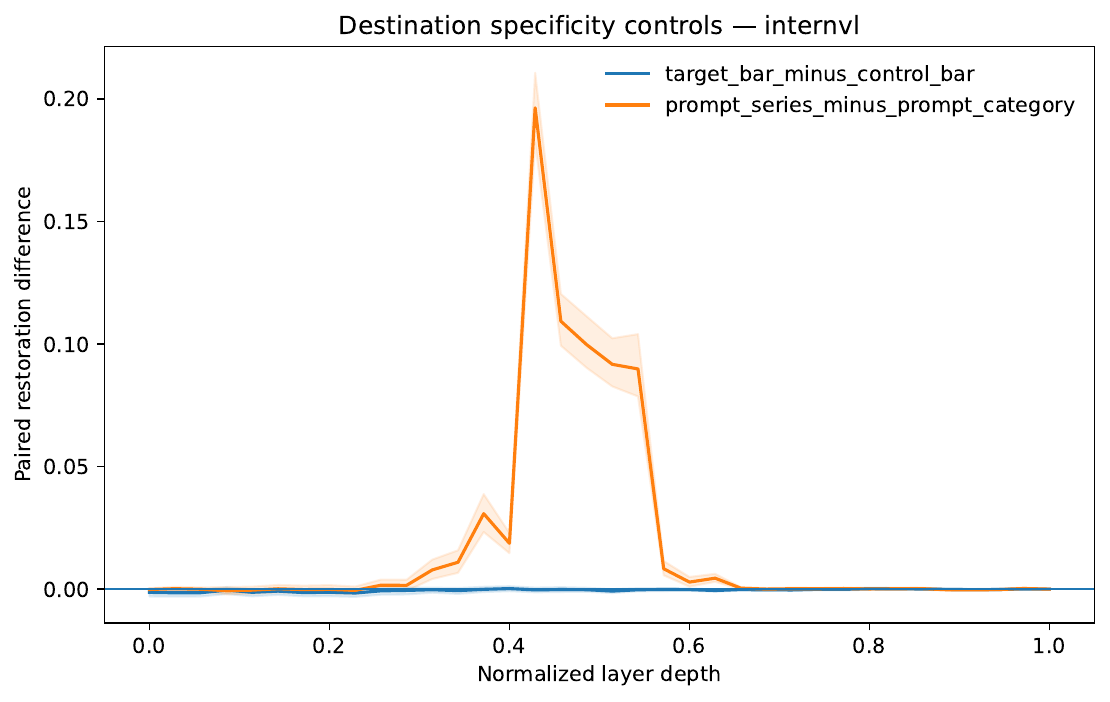}
    \includegraphics[width=0.49\textwidth]{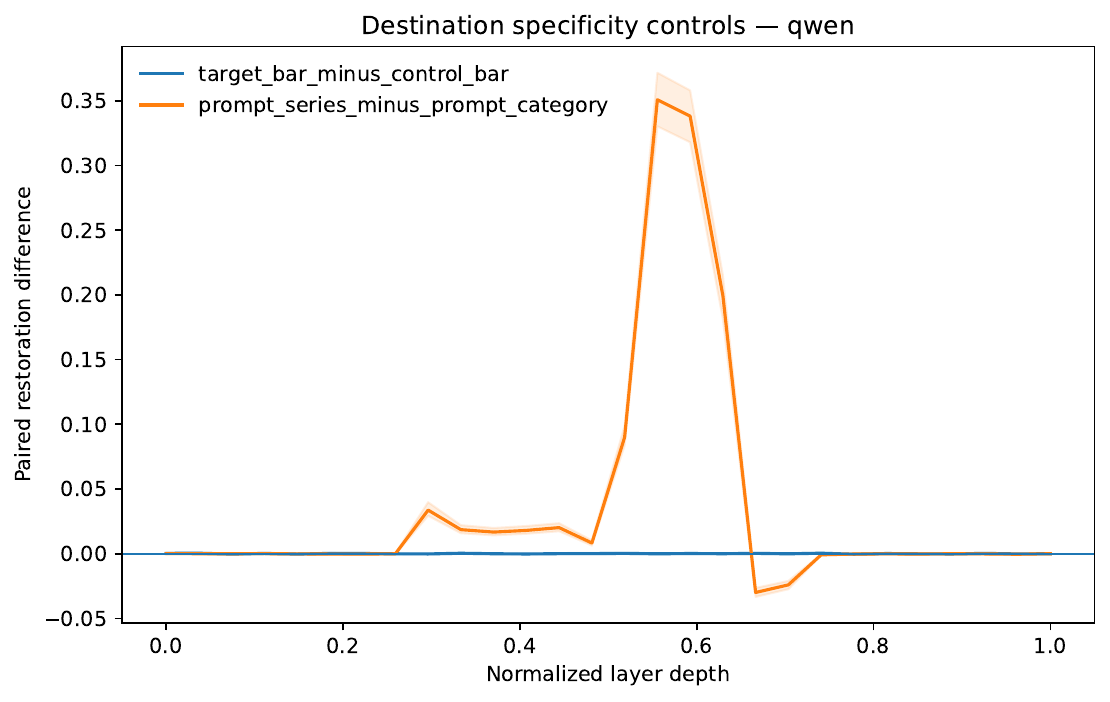}
    \caption{Layer-wise differences between each candidate destination and its matched control for InternVL3.5-8B (left) and Qwen2.5-VL-7B (right). Shaded bands denote pointwise 95\% confidence intervals from 500 query-level bootstrap resamples.}
    \label{fig:legend_destination_specificity}
\end{figure*}

\paragraph{The prompt-series effect is stronger than the matched controls.}
Figure~\ref{fig:legend_destination_specificity} compares the candidate locations with their corresponding controls. The prompt-series positions show stronger restoration than the prompt-label positions, with the difference concentrated in middle layers. By contrast, restoration from the target-bar region remains close to that of the matched control-bar region. These results identify middle-layer prompt-series states as candidates with a local causal effect on answer preference under legend swaps. Nevertheless, independently restoring these states does not establish whether they contribute to the effect of an earlier visual-legend intervention. We therefore conduct a second experiment to test whether resetting the prompt-series states reduces the restoration produced by an earlier visual-legend patch.

\subsubsection{Testing Casual dependence Experiment Setup Details}
\label{sec:testingcasualdependencydetails}
We next test whether middle-layer prompt-series states contribute to the restoration effect produced by an earlier visual-legend patch. To examine this relationship, we first restore the visual-legend states and then reset the prompt-series states at a deeper layer.

\paragraph{Source restoration.}

Using the counterfactual input as the receiver computation, we replace the hidden states at the image-token positions corresponding to the two affected visual-legend entries with their original-input counterparts at a selected language-model layer.  We refer to the resulting normalized restoration $R$ as \emph{source rescue}.

\paragraph{Layer selection on the pilot set.}

For each model, we first scan all language-model layers on the 100 pilot queries and select the source layer that produces the largest mean restoration from the affected visual-legend positions. After fixing the source layer, we scan deeper candidate layers and reset the prompt-series positions to their counterfactual hidden states. The layer producing the largest reduction in source rescue is selected as the reset layer. Both layers are then fixed before evaluation on the held-out set.

\paragraph{Effect estimation on the held-out set.}
On the 400 held-out queries, we first restore the original legend-entry activations at the selected source layer to obtain the unblocked source-rescue baseline. We then reset the prompt-series positions to their counterfactual activations at the selected destination layer and measure the resulting decrease in source rescue. As controls, we separately reset the prompt-label positions, the target-bar region, and a matched control-bar region at the same destination layer. The control-bar region is constructed at a randomly selected non-query category. These controls test whether the observed decrease is specific to the prompt-series positions. All blocking effects and their 95\% confidence intervals are estimated on the held-out set using 5,000 source-chart cluster-bootstrap resamples (Eq~\ref{eq:app_cluster_bootstrap}).

On the 400 held-out queries, we first restore the original visual-legend states at the selected source layer to obtain the source-rescue baseline. We then reset the prompt-series positions to their hidden states from the unpatched counterfactual computation at the selected deeper layer and measure the resulting change in restoration.

As controls, we separately reset the prompt-label positions, the target-bar region, and a matched control-bar region at the same reset layer. The control-bar region is constructed at a randomly selected non-query category. These controls allow us to assess whether the reduction in source rescue is specific to the prompt-series positions among the tested locations.

For each reset location $d$, we define the reduction in source rescue as
\begin{equation}
\Delta R_d =
R_{\mathrm{source\ only}} -
R_{\mathrm{source+reset}(d)}.
\label{eq:source_reset_effect}
\end{equation}
A positive $\Delta R_d$ indicates that resetting location $d$ reduces the restoration produced by the earlier visual-legend patch. We estimate the reset effects and their 95\% confidence intervals using 5,000 source-chart cluster-bootstrap resamples on the held-out set.

\subsection{Factorial Recombination Analysis Evaluation Metrics Details}
\label{sec:fraemd}

For each family, the model scores the four candidate values as complete answer sequences. For a candidate answer $c=(c_1,\ldots,c_T)$, its sequence score under condition $x$ is

\begin{equation}
S_x(c)
=
\sum_{t=1}^{T}
\log P\!\left(c_t \mid x,q,c_{<t}\right),
\label{eq:factorial-sequence-score}
\end{equation}

where $q$ is the fixed prompt and $x$ specifies the input image and, when applicable, the intervention. The score is the sum of the conditional log-probabilities of all answer tokens.

\subsubsection{Four-Candidate Target Accuracy}

At the projected image-token site, we report four-candidate target accuracy for all four interventions. Each intervention has a predefined target answer: the correct answer of the geometry donor for geometry-only, the correct answer of the scale donor for scale-only, and the correct answer of the diagonal for both joint-cross and joint-coherent.

A family is counted as a hit if the predefined target has a strictly higher sequence score than each of the other three candidates. Ties and lower-ranked targets are counted as misses. Target accuracy is the mean hit rate across families. This metric measures answer selection among the four predefined candidates; it is not free-generation accuracy.

\subsubsection{Margins and Normalized Restoration}

To match the endpoint notation used in the preceding analyses, we denote the receiver input by counterfactual (\textnormal{cf}) and the diagonal input by original (\textnormal{ori}). Both use the same prompt. Let $c_{\mathrm{cf}}$ be the correct answer for the receiver and $c_{\mathrm{ori}}$ the correct answer for the diagonal. The two unpatched baseline margins are

\begin{align}
M_{\mathrm{cf}}
&=
S_{\mathrm{cf}}(c_{\mathrm{ori}})
-
S_{\mathrm{cf}}(c_{\mathrm{cf}}),
\label{eq:factorial-margin-cf}
\\
M_{\mathrm{ori}}
&=
S_{\mathrm{ori}}(c_{\mathrm{ori}})
-
S_{\mathrm{ori}}(c_{\mathrm{cf}}).
\label{eq:factorial-margin-ori}
\end{align}

Thus, $M_{\mathrm{cf}}$ measures the model's preference for $c_{\mathrm{ori}}$ over $c_{\mathrm{cf}}$ when the receiver is the input, whereas $M_{\mathrm{ori}}$ measures the same preference when the diagonal is the input.

We define a separate patched margin for each joint intervention:

\begin{align}
M_{\mathrm{patch}}^{\mathrm{joint\text{-}cross}}
&=
S_{\mathrm{patch}}^{\mathrm{joint\text{-}cross}}(c_{\mathrm{ori}})
\nonumber\\
&\quad-
S_{\mathrm{patch}}^{\mathrm{joint\text{-}cross}}(c_{\mathrm{cf}}),
\label{eq:factorial-margin-cross}
\\
M_{\mathrm{patch}}^{\mathrm{joint\text{-}coherent}}
&=
S_{\mathrm{patch}}^{\mathrm{joint\text{-}coherent}}(c_{\mathrm{ori}})
\nonumber\\
&\quad-
S_{\mathrm{patch}}^{\mathrm{joint\text{-}coherent}}(c_{\mathrm{cf}}).
\label{eq:factorial-margin-coherent}
\end{align}

The first margin is obtained by patching the target-bar region from the geometry donor and the y-axis scale region from the scale donor into the receiver forward pass. The second is obtained by patching both regions from the diagonal into the receiver forward pass.

The corresponding normalized restoration scores are

\begin{align}
R_{\mathrm{joint\text{-}cross}}
&=
\frac{
M_{\mathrm{patch}}^{\mathrm{joint\text{-}cross}}-M_{\mathrm{cf}}
}{
M_{\mathrm{ori}}-M_{\mathrm{cf}}
},
\label{eq:factorial-restoration-cross}
\\
R_{\mathrm{joint\text{-}coherent}}
&=
\frac{
M_{\mathrm{patch}}^{\mathrm{joint\text{-}coherent}}-M_{\mathrm{cf}}
}{
M_{\mathrm{ori}}-M_{\mathrm{cf}}
}.
\label{eq:factorial-restoration-coherent}
\end{align}

For either joint condition, $R=0$ means that the intervention leaves the answer preference at the receiver (cf) baseline, whereas $R=1$ means that it restores the full preference difference between the receiver (cf) and diagonal (ori) baselines. 

\subsubsection{Statistical Analysis Metrics}

For the projected image-token site, 95\% confidence intervals for four-candidate target accuracy, $R_{\mathrm{joint\text{-}cross}}$, and $R_{\mathrm{joint\text{-}coherent}}$ are computed from 5,000 family-level bootstrap resamples.

The layer-wise analysis reports separate curves for $R_{\mathrm{joint\text{-}cross}}$ and $R_{\mathrm{joint\text{-}coherent}}$; pointwise 95\% confidence intervals use 500 family-level bootstrap resamples. In every analysis, the complete family is the sampling unit.

\begin{figure*}[t]
    \centering
    \includegraphics[width=\textwidth]{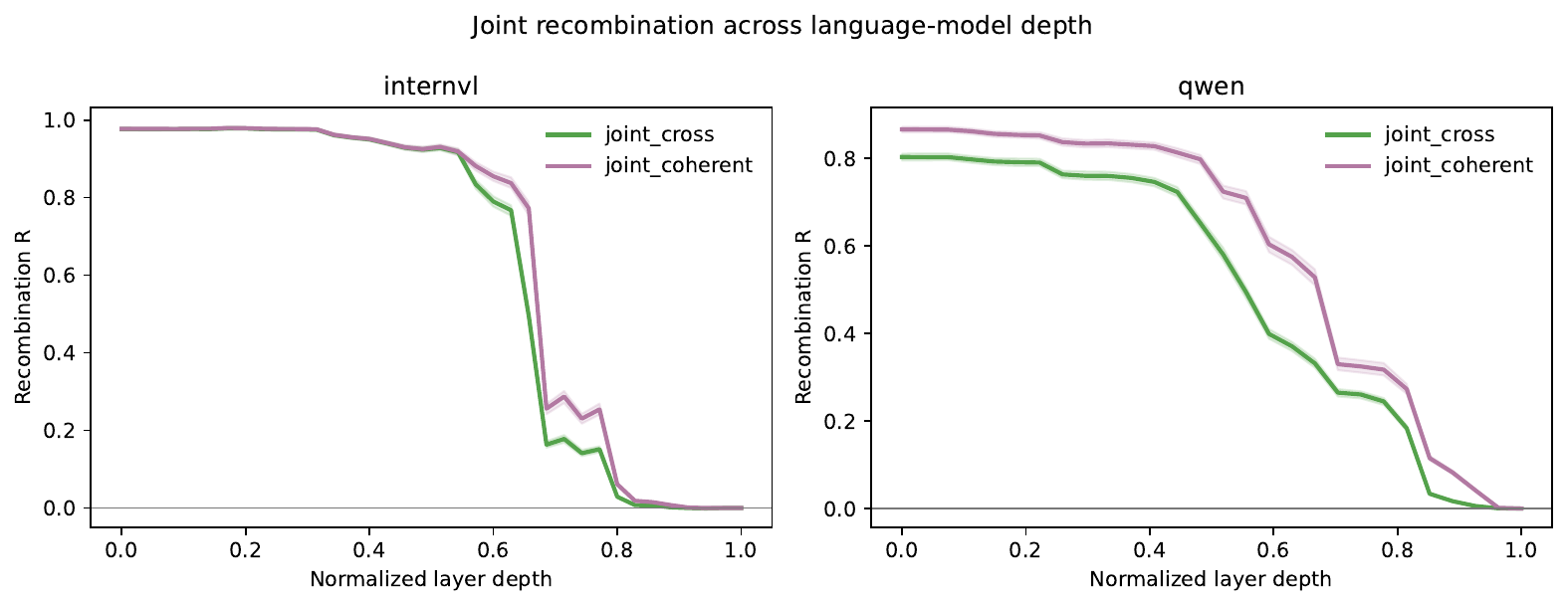}
    \caption{Layer-wise normalized restoration for joint-cross and joint-coherent. Shading shows pointwise 95\% confidence intervals from 500 family-level bootstrap resamples.}
    \label{fig:factorial-layerwise-restoration}
\end{figure*}

\clearpage

\subsection{Chart Value Reading Data Generation Details and Evaluation details (single-factor analysis and casual role of prompt-series state)}
\label{sec:chartvaluereadingdatagendetail}

\begin{table*}[t]
\centering

\scriptsize
\setlength{\tabcolsep}{2.8pt}
\renewcommand{\arraystretch}{0.96}
\resizebox{\textwidth}{!}{
\begin{tabular}{
    lr lr lr lr
    @{\hspace{6pt}}
    lr lr
}
\toprule
\multicolumn{8}{c}{
    X-label pseudowords
    (2,493 chart-level occurrences)
}
&
\multicolumn{4}{c}{
    Series/legend pseudowords
    (1,150 chart-level occurrences)
}
\\
\cmidrule(lr){1-8}
\cmidrule(lr){9-12}
Token & Count &
Token & Count &
Token & Count &
Token & Count &
Token & Count &
Token & Count
\\
\midrule
\texttt{Bebe} & 61 &
\texttt{Beha} & 42 &
\texttt{Beja} & 55 &
\texttt{Cag}  & 52 &
\texttt{Bahe} & 56 &
\texttt{Boz}  & 43
\\

\texttt{Cib}  & 40 &
\texttt{Ciha} & 46 &
\texttt{Daj}  & 43 &
\texttt{Dega} & 54 &
\texttt{Ceh}  & 48 &
\texttt{Cul}  & 37
\\

\texttt{Deje} & 59 &
\texttt{Fece} & 41 &
\texttt{Fehe} & 38 &
\texttt{Fiha} & 51 &
\texttt{Dac}  & 46 &
\texttt{Dag}  & 49
\\

\texttt{Gaz}  & 60 &
\texttt{Gor}  & 49 &
\texttt{Guh}  & 48 &
\texttt{Haj}  & 57 &
\texttt{Fafa} & 42 &
\texttt{Fup}  & 38
\\

\texttt{Hav}  & 47 &
\texttt{Higa} & 43 &
\texttt{Jide} & 55 &
\texttt{Jof}  & 53 &
\texttt{Geh}  & 56 &
\texttt{Gibe} & 43
\\

\texttt{Jov}  & 57 &
\texttt{Kede} & 42 &
\texttt{Kice} & 36 &
\texttt{Kike} & 54 &
\texttt{Haf}  & 39 &
\texttt{Hece} & 54
\\

\texttt{Lad}  & 46 &
\texttt{Lede} & 51 &
\texttt{Lefe} & 52 &
\texttt{Maga} & 55 &
\texttt{Jec}  & 45 &
\texttt{Juz}  & 42
\\

\texttt{Mece} & 51 &
\texttt{Meka} & 34 &
\texttt{Ner}  & 34 &
\texttt{Nica} & 60 &
\texttt{Kul}  & 52 &
\texttt{Lica} & 46
\\

\texttt{Nin}  & 51 &
\texttt{Peb}  & 56 &
\texttt{Peka} & 57 &
\texttt{Pir}  & 40 &
\texttt{Moc}  & 50 &
\texttt{Nige} & 60
\\

\texttt{Reca} & 51 &
\texttt{Rof}  & 48 &
\texttt{Rug}  & 41 &
\texttt{Sabe} & 65 &
\texttt{Puj}  & 54 &
\texttt{Ruc}  & 39
\\

\texttt{Sebe} & 53 &
\texttt{Suf}  & 47 &
\texttt{Tas}  & 40 &
\texttt{Tig}  & 48 &
\texttt{Seb}  & 54 &
\texttt{Tiga} & 51
\\

\texttt{Tum}  & 43 &
\texttt{Vig}  & 156 &
\texttt{Zida} & 62 &
\texttt{Zif}  & 69 &
\texttt{Vije} & 49 &
\texttt{Zede} & 57
\\
\bottomrule
\end{tabular}
}
\caption{
    Chart-level occurrence counts of the pseudowords used in the
    500 unique source charts.
    We report counts at the source-chart level rather than the
    5,932-query level because each chart is repeated once for every
    target bar, which would otherwise overweight charts containing
    more bars.
    The x-label vocabulary contains 48 pseudowords and the
    series/legend vocabulary contains 24 pseudowords.
}

\label{tab:pseudoword_distribution}

\end{table*}

\begin{figure*}[t]
    \centering
    \includegraphics[
        width=\textwidth
    ]{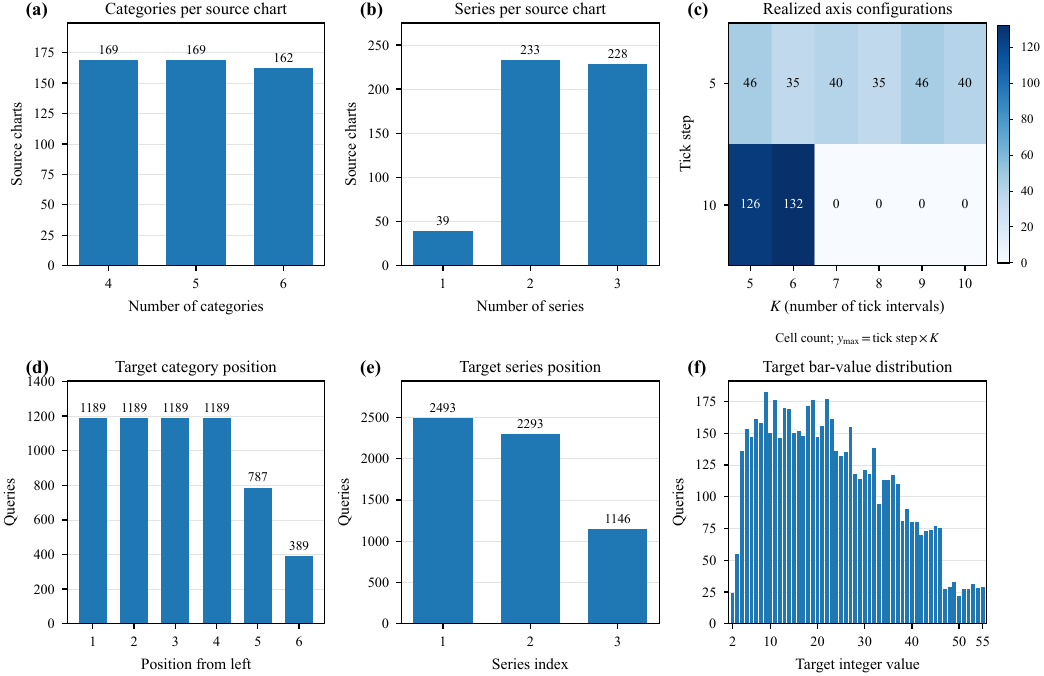}
    \caption{
        Realized distribution of the controlled dataset.
        Panels (a)--(c) aggregate the 500 unique source charts,
        whereas panels (d)--(f) aggregate the 5,932 read-value
        queries derived from these charts.
        Panel (c) reports the number of source charts for each
        realized pair of tick step and number of tick intervals,
        where $y_{\max}=\text{tick step}\times K$.
        The lower frequencies at category positions 5--6 and
        series positions 2--3 reflect structural availability:
        not every chart contains six categories or three series.
    }
    \label{fig:chartvaluereadingdatadis}
\end{figure*}

\paragraph{Intial Chart Reading Data distribution}
We show the distributions of the 500 synthesized charts and 5932 chart value reading samples in Table~\ref{tab:pseudoword_distribution} and Figure~\ref{fig:chartvaluereadingdatadis}. The code templates for the charts are shown in Figure~\ref{fig:bar-generation-template}.

\paragraph{Counterfactual Chart Generation}
Complete counterfactual construction and patched-support definitions for all nine analysis conditions are provided in Table~\ref{tab:interventions}.

\paragraph{Evaluation Results on Original and Counterfactal Chart Value Reading Full results}
\label{sec:full_res_value_reading}

\begin{table*}[t]
\centering
\scriptsize
\setlength{\tabcolsep}{3.4pt}
\renewcommand{\arraystretch}{1.12}
\begin{tabular}{@{}lllrrrrrrrr@{}}
\toprule
Model
& Scope
& Counterfactual
& FF
& FT
& TF
& TT
& Ori.\ Acc.
& CF Acc.
& Both
& Consistency \\
\midrule

\multirow{7}{*}{Qwen2.5-VL-7B}
& \multirow{4}{*}{Target}
& Axis scale
& 142 & 2 & 1516 & 4272
& 97.57 & 72.05 & 72.02 & -- \\

&
& Bar height
& 10 & 134 & 134 & 5654
& 97.57 & 97.57 & 95.31 & -- \\

&
& Legend swap
& 13 & 131 & 140 & 5648
& 97.57 & 97.42 & 95.21 & -- \\

&
& X-label swap
& 12 & 132 & 122 & 5666
& 97.57 & 97.74 & 95.52 & -- \\

\cmidrule(lr){2-11}

&
\multirow{3}{*}{Non-target}
& Bar height
& 124 & 20 & 14 & 5774
& 97.57 & 97.67 & 97.34 & 99.43 \\

&
& X-label swap
& 137 & 7 & 7 & 5781
& 97.57 & 97.57 & 97.45 & 99.76 \\

&
& Legend swap
& 141 & 3 & 3 & 5785
& 97.57 & 97.57 & 97.52 & 99.90 \\

\midrule

\multirow{7}{*}{InternVL3.5-8B}
& \multirow{4}{*}{Target}
& Axis scale
& 145 & 17 & 731 & 5039
& 97.27 & 85.23 & 84.95 & -- \\

&
& Bar height
& 70 & 92 & 108 & 5662
& 97.27 & 97.00 & 95.45 & -- \\

&
& Legend swap
& 33 & 129 & 133 & 5637
& 97.27 & 97.20 & 95.03 & -- \\

&
& X-label swap
& 43 & 119 & 131 & 5639
& 97.27 & 97.07 & 95.06 & -- \\

\cmidrule(lr){2-11}

&
\multirow{3}{*}{Non-target}
& Bar height
& 143 & 19 & 19 & 5751
& 97.27 & 97.27 & 96.95 & 99.36 \\

&
& X-label swap
& 141 & 21 & 15 & 5755
& 97.27 & 97.37 & 97.02 & 99.39 \\

&
& Legend swap
& 143 & 19 & 15 & 5755
& 97.27 & 97.34 & 97.02 & 99.43 \\

\bottomrule
\end{tabular}

\caption{
Behavioral transitions between the original and image-counterfactual
charts over \(N=5{,}932\) fixed-prompt pairs per condition.
F and T denote strict incorrectness and correctness, respectively, on
the original and counterfactual endpoints; thus, TF denotes an
originally correct but counterfactually incorrect pair.
Ori.\ Acc.\ and CF Acc.\ are the endpoint accuracies.
Both is \(N_{\mathrm{TT}}/N\).
For non-target edits, Consistency is
\((N_{\mathrm{FF}}+N_{\mathrm{TT}})/N\), measuring whether strict
correctness is preserved after an edit that leaves the gold answer
unchanged. Percentages are reported in the final four columns.
}
\label{tab:cf_behavior}
\end{table*}

The 500 source charts yield 5,932 value-reading
queries, and each counterfactual is evaluated using the same prompt on
the original and edited charts. A \emph{target-relevant} edit changes
visual evidence that determines the queried value. A
\emph{non-target} edit instead applies an analogous operation away from
the queried bar: changing another bar, swapping two category labels
that exclude the target category, or swapping two legend mappings that
exclude the target series. The gold answer therefore remains unchanged
under a non-target edit. We do not construct a non-target axis-scale
condition because the axis scale globally affects the numerical
interpretation of all bars.
On the original charts, strict accuracy is \(97.57\%\) for
Qwen2.5-VL-7B and \(97.27\%\) for InternVL3.5-8B. For target-relevant
bar-height, legend-swap, and x-label-swap counterfactuals, both endpoints
are answered correctly for \(95.03\%\)--\(95.52\%\) of the pairs. Axis
scaling is more difficult, with both-endpoint correctness of \(72.02\%\)
for Qwen2.5-VL-7B and \(84.95\%\) for InternVL3.5-8B. For non-target
edits, the strict-correctness status is preserved on
\(99.36\%\)--\(99.90\%\) of the pairs, indicating that local edits
unrelated to the queried bar rarely change whether the answer is
correct.

Table~\ref{tab:cf_behavior} reports the complete strict-correctness
transitions between the original and counterfactual charts. We denote
each pair by its correctness state on the original and counterfactual
endpoints: for example, TF indicates that the model is strictly correct
on the original chart but incorrect on the counterfactual chart.
For target-relevant edits, \emph{Both} is the proportion of pairs for
which both endpoints are answered correctly, \(N_{\mathrm{TT}}/N\).
For non-target edits, whose gold answer remains unchanged, we
additionally report correctness consistency,
\((N_{\mathrm{FF}}+N_{\mathrm{TT}})/N\).

Legend edits have additional feasibility constraints. A target-relevant
legend swap requires at least two series: 5,732 of the 5,932 queries
admit an actual swap, while 200 single-series queries are no-op pairs.
A non-target legend swap requires two series other than the target:
3,438 queries from three-series charts admit an actual edit, while the
remaining 2,494 pairs are pixel-identical no-ops. The aggregate legend
rows in Table~\ref{tab:cf_behavior} include these no-op pairs and should
therefore be interpreted with this qualification. For correctly
answered pixel-identical pairs, the absolute change in the gold-answer
log probability satisfies
\(\lvert \Delta \log p(y^\star) \rvert < 10^{-3}\).

\begin{figure*}[t]
\centering

\begin{minipage}[t]{0.475\textwidth}
\vspace{0pt}
\textbf{(a) Parameter sampling}
\begin{lstlisting}[style=barcompact]
# Chart structure
num_series    = sample({1, 2, 3})
num_categories = sample({4, 5, 6})

# Axis scale
tick_step = sample({5, 10})
num_ticks = sample({5, ..., 10})
y_max = tick_step * num_ticks
require y_max <= 60

# Bar values leave space for editing
v_min = ceil(0.08 * y_max)
v_max = floor(0.92 * y_max)
values = sample_integers(
    shape=(num_series, num_categories),
    range=[v_min, v_max])

# Visual attributes
labels = sample_unique_pseudowords()
colors = sample_distinct_colors()

legend = None if num_series == 1 \
    else sample_fixed_legend_layout()

# Same chart at two resolutions
outputs = [(448, 64), (1344, 192)]
\end{lstlisting}
\end{minipage}
\hfill
\begin{minipage}[t]{0.475\textwidth}
\vspace{0pt}
\textbf{(b) Fixed-layout rendering}
\begin{lstlisting}[style=barcompact]
fig = plt.figure(figsize=(7, 7), dpi=dpi)
ax = fig.add_axes([.125, .165, .835, .700])

for s in range(num_series):
    xpos = grouped_bar_positions(
        s, num_series, num_categories)
    ax.bar(xpos, values[s],
           width=group_width / num_series,
           color=colors[s],
           edgecolor="none")

ax.set_xlim(-0.6, num_categories - 0.4)
ax.set_ylim(0, y_max)
ax.set_xticks(category_positions)
ax.set_yticks(range(0, y_max + 1, tick_step))

# Replace automatic text placement
ax.set_xticklabels([])
ax.set_yticklabels([])
draw_ticks_at_fixed_slots(ax, labels)
draw_legend_at_fixed_slot(fig, legend)

ax.grid(False)
ax.minorticks_off()

# Preserve the fixed canvas geometry
fig.savefig(path, dpi=dpi,
            bbox_inches=None,
            pad_inches=0)
\end{lstlisting}
\end{minipage}

\caption{
\textbf{Controlled vertical-bar chart generation template.}
Chart structure, integer bar values, labels, colors, axis scale,
and legend configuration are sampled explicitly. Rendering uses
a fixed canvas, axes rectangle, grouped-bar geometry, and
predefined text slots. The paired low- and high-resolution images
share all semantic and normalized layout parameters and differ
only in DPI. Automatic tick-label placement, automatic legends,
tight layout, and tight bounding-box cropping are disabled.
}
\label{fig:bar-generation-template}
\end{figure*}

\subsection{Evaluation Data Audit  (single-factor analysis and casual role of prompt-series state)}
\label{app:eval_data_audit}

This appendix reports how the activation-patching evaluation set is formed
and summarizes its realized coverage. The sampler uses seed 42 and returns
500 queries for each of the nine analysis conditions, giving 4,500 records
per model. The visual conditions form 500 complete bundles and the prompt
conditions form another 500 complete bundles. Counting each pool-level target
once gives 1,500 selections and 1,361 distinct target-bar IDs from 464 of the
500 source charts. Within every condition, the two models use identical
sample IDs, order, and intervention metadata. All selected pairs have
different original and counterfactual gold values, and both models answer
both endpoints exactly.

\subsubsection{Selection rule and evaluation scope}

A pair is \emph{cross-model eligible} when both models answer both the
original and counterfactual endpoints exactly and the two gold values differ.
This synchronized-success rule ensures that activation restoration is measured
on examples for which both endpoints are solved by both models. The resulting
analysis characterizes successful value-reading computations; the composition
checks below verify that this subset retains broad coverage over the measured
chart and target factors.

Table~\ref{tab:app_sample_funnel} reports the image-counterfactual selection
funnel. Axis scaling is the most selective condition, while the three local
image edits retain approximately 89--91\% of feasible pairs. The primary
bar-region comparison is especially well controlled: bar union, difference,
and intersection use the same 500 original--counterfactual pairs and differ
only in the patched ROI. Thus, their paired contrast does not mix different
success-selected targets.

\begin{table*}[t]
\centering
\small
\setlength{\tabcolsep}{5.2pt}
\renewcommand{\arraystretch}{1.12}
\resizebox{\textwidth}{!}{%
\begin{tabular}{@{}lrrrrrr@{}}
\toprule
Counterfactual
& Initial pairs
& Feasible pairs
& \qwen{} both correct
& \intern{} both correct
& Cross-model eligible
& Final sampled \\
\midrule
Axis scale
& 5,932 & 5,932 & 4,272 & 5,039 & 3,720 (62.7\%) & 500 \\
Bar height
& 5,932 & 5,932 & 5,654 & 5,662 & 5,408 (91.2\%) & 500 \\
Legend swap
& 5,932 & 5,732 & 5,648 & 5,637 & 5,121 (89.3\%) & 500 \\
X-label swap
& 5,932 & 5,932 & 5,666 & 5,639 & 5,397 (91.0\%) & 500 \\
\bottomrule
\end{tabular}%
}
\caption{
Selection funnel for image counterfactuals. ``Both correct'' means that one
model answers both endpoints exactly. Cross-model eligibility requires all
four endpoint predictions to be correct and the two gold values to differ.
A target-relevant legend swap requires at least two series, so 200
single-series queries are excluded from its feasible denominator. The
model-level both-correct columns are measured on all 5,932 generated pairs.
}
\label{tab:app_sample_funnel}
\end{table*}

\subsubsection{Evaluation pools and shared units}

A bundle contains conditions evaluated on the same original query sources,
which enables within-query comparisons. Table~\ref{tab:app_bundle_definition}
states what is shared in each pool.

\begin{table*}[t]
\centering
\small
\setlength{\tabcolsep}{4.5pt}
\renewcommand{\arraystretch}{1.12}
\begin{tabularx}{\textwidth}{@{}lp{2.2cm}X@{}}
\toprule
Pool & Conditions & Shared source relation \\
\midrule
Visual bundle
& Bar union, bar difference, bar intersection, legend swap, x-label swap
& All five conditions use the same 500 original query IDs and target bars
from 417 source charts. The three bar conditions additionally use the exact
same original--counterfactual bar-height pairs and differ only in the ROI.
Legend and x-label interventions apply their own edits to the same original
query sources. \\

Prompt bundle
& Label, series, label+series
& All three conditions use the same 500 original images, query IDs, and
target bars from 418 source charts. Each condition applies its own prompt
edit while keeping the image fixed. \\

Axis pool
& Axis scale
& Axis scale uses a separately sampled set of 500 query IDs from 307 source
charts. \\
\bottomrule
\end{tabularx}
\caption{
Definition of the three evaluation pools. The two models use identical
sample IDs, order, and intervention metadata within every condition.
}
\label{tab:app_bundle_definition}
\end{table*}

Table~\ref{tab:app_nine_types} lists the nine analysis conditions. The three
bar rows share one bar-height counterfactual and differ only in the ROI.
Prompt conditions patch text positions and therefore have no image ROI.

\begin{table*}[t]
\centering
\small
\setlength{\tabcolsep}{4.5pt}
\renewcommand{\arraystretch}{1.10}
\resizebox{\textwidth}{!}{%
\begin{tabular}{@{}llrrrrrr@{}}
\toprule
Modality & Condition & Queries & Charts & Target bars
& \intern{} ROI mean & \qwen{} ROI mean & Pool \\
\midrule
\multirow{6}{*}{Image}
& Axis scale       & 500 & 307 & 500 & 3.048\% & 2.309\% & Axis \\
& Bar union        & 500 & 417 & 500 & 2.462\% & 2.106\% & Visual \\
& Bar difference   & 500 & 417 & 500 & 0.667\% & 0.581\% & Visual \\
& Bar intersection & 500 & 417 & 500 & 1.795\% & 1.524\% & Visual \\
& Legend swap      & 500 & 417 & 500 & 0.918\% & 0.684\% & Visual \\
& X-label swap     & 500 & 417 & 500 & 1.336\% & 1.122\% & Visual \\
\midrule
\multirow{3}{*}{Prompt}
& Label            & 500 & 418 & 500 & -- & -- & Prompt \\
& Series           & 500 & 418 & 500 & -- & -- & Prompt \\
& Label + series   & 500 & 418 & 500 & -- & -- & Prompt \\
\bottomrule
\end{tabular}%
}
\caption{
Composition of the nine conditions. ROI values are mean fractions of the
rendered image area; raw pixel areas are not compared across resolutions.
}
\label{tab:app_nine_types}
\end{table*}

The visual bundle contains 334 charts used once and 83 used twice. The prompt
bundle contains 336 charts used once and 82 used twice. The axis pool contains
176 charts used once, 91 used twice, 26 used three times, 11 used four times,
one used six times, and two used seven times. Among the 1,500 pool-level
target selections, 1,361 target-bar IDs are distinct, so some targets recur
across pools. At the source-chart level, the union contains 464 charts, and
the pairwise overlaps are reported in Table~\ref{tab:app_pool_overlap}. The
main statistical analysis therefore includes source-chart cluster bootstrap
intervals and chart-equal estimates in addition to query-level estimates.

\begin{table}[t]
\centering
\scriptsize
\setlength{\tabcolsep}{3.0pt}
\begin{tabular}{@{}lrrrr@{}}
\toprule
Pool & Queries & Charts & Mean/chart & Max/chart \\
\midrule
Visual bundle & 500 & 417 & 1.20 & 2 \\
Prompt bundle & 500 & 418 & 1.20 & 2 \\
Axis scale    & 500 & 307 & 1.63 & 7 \\
\bottomrule
\end{tabular}
\caption{Query and source-chart counts in the three evaluation pools.}
\label{tab:app_pool_coverage}
\end{table}

\begin{table}[t]
\centering
\small
\begin{tabular}{@{}lrrr@{}}
\toprule
 & Axis & Visual & Prompt \\
\midrule
Axis   & 307 & 270 & 275 \\
Visual & 270 & 417 & 389 \\
Prompt & 275 & 389 & 418 \\
\bottomrule
\end{tabular}
\caption{
Source-chart overlap among the three pools. Diagonal entries give the number
of charts in each pool.
}
\label{tab:app_pool_overlap}
\end{table}

\subsubsection{Target-value and position coverage}

\begin{figure*}[t]
\centering
\includegraphics[width=\textwidth]{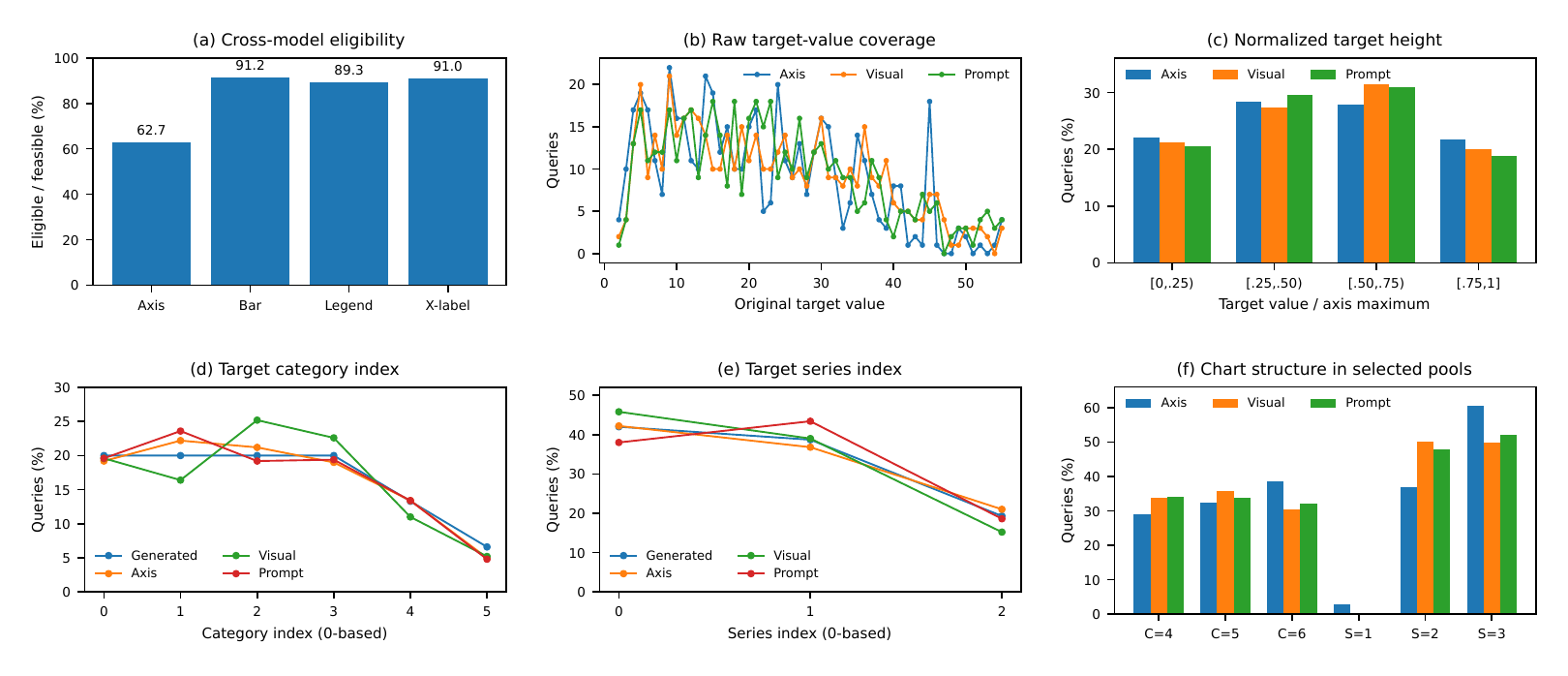}
\caption{
Composition of the evaluation pools. (a) Cross-model eligibility among
feasible image pairs. (b) Exact original target-value frequencies.
(c) Target height normalized by the original axis maximum. (d--e) Exact
target category and series positions, including the complete generated pool
as a reference. (f) Realized numbers of categories (\(C\)) and series
(\(S\)). Panels (b)--(f) count each bundle once rather than repeating the
same target for every condition.
}
\label{fig:app_eval_distribution}
\end{figure*}

Figure~\ref{fig:app_eval_distribution} summarizes the main distribution
checks. Across the three selected pools, original target values span 2--55,
and normalized target heights span 8--92\% of the axis maximum. The pool
means are also similar: 21.23--22.83 for the raw value and 48.84--49.37\%
for normalized height. Each of the four normalized-height bins contains at
least 18.8\% of every pool. The selected pools retain all target category and
series indices available in the generated 5,932-query set. Later indices are
less frequent because they occur only in charts with more categories or
series, a pattern already present in the generated pool.

\begin{table*}[t]
\centering
\small
\setlength{\tabcolsep}{4.2pt}
\renewcommand{\arraystretch}{1.08}
\resizebox{\textwidth}{!}{%
\begin{tabular}{@{}lrrrrrrrrrrrr@{}}
\toprule
& \multicolumn{6}{c}{Original target value}
& \multicolumn{6}{c}{Normalized target height (\%)} \\
\cmidrule(lr){2-7}\cmidrule(lr){8-13}
Pool
& Mean & SD & Min & Q1 & Median & Q3--Max
& Mean & SD & Min & Q1 & Median & Q3--Max \\
\midrule
Axis
& 21.23 & 12.54 & 2 & 11 & 20 & 30--55
& 48.84 & 25.38 & 8.00 & 25.71 & 48.45 & 70.00--92.00 \\
Visual
& 22.79 & 12.98 & 2 & 12 & 21 & 33--55
& 49.10 & 23.99 & 8.00 & 27.88 & 50.00 & 70.00--92.00 \\
Prompt
& 22.83 & 12.90 & 2 & 12 & 21 & 31--55
& 49.37 & 24.37 & 8.00 & 30.00 & 48.73 & 70.28--92.00 \\
\bottomrule
\end{tabular}%
}
\caption{
Target-value summaries. Normalized height is the original target value
divided by the original chart's axis maximum. Each visual or prompt bundle
is counted once.
}
\label{tab:app_target_value_summary}
\end{table*}

\begin{table*}[t]
\centering
\scriptsize
\setlength{\tabcolsep}{3.4pt}
\begin{tabular}{@{}lrrrrrr|rrr@{}}
\toprule
& \multicolumn{6}{c|}{Target category index (\%)}
& \multicolumn{3}{c}{Target series index (\%)} \\
\cmidrule(lr){2-7}\cmidrule(lr){8-10}
Pool & 0 & 1 & 2 & 3 & 4 & 5 & 0 & 1 & 2 \\
\midrule
Generated (5,932) & 20.0 & 20.0 & 20.0 & 20.0 & 13.3 & 6.6
                  & 42.0 & 38.7 & 19.3 \\
Axis (500)        & 19.2 & 22.2 & 21.2 & 19.0 & 13.4 & 5.0
                  & 42.2 & 36.8 & 21.0 \\
Visual (500)      & 19.6 & 16.4 & 25.2 & 22.6 & 11.0 & 5.2
                  & 45.8 & 39.0 & 15.2 \\
Prompt (500)      & 19.6 & 23.6 & 19.2 & 19.4 & 13.4 & 4.8
                  & 38.0 & 43.4 & 18.6 \\
\bottomrule
\end{tabular}
\caption{
Exact target-position percentages. The generated-pool row uses all 5,932
queries; the three evaluation rows count one representative target per pool.
Later indices are structurally available only in charts with enough
categories or series.
}
\label{tab:app_exact_target_indices}
\end{table*}

All three selected pools cover four-, five-, and six-category charts. The
visual and prompt bundles require at least two series, whereas the axis pool
can include single-series charts.

\begin{table*}[t]
\centering
\small
\setlength{\tabcolsep}{5pt}
\begin{tabular}{@{}lrrrrrr@{}}
\toprule
& \multicolumn{3}{c}{Number of categories}
& \multicolumn{3}{c}{Number of series} \\
\cmidrule(lr){2-4}\cmidrule(lr){5-7}
Pool & 4 & 5 & 6 & 1 & 2 & 3 \\
\midrule
Axis scale    & 145 & 162 & 193 & 14 & 184 & 302 \\
Visual bundle & 169 & 179 & 152 & 0  & 251 & 249 \\
Prompt bundle & 171 & 169 & 160 & 0  & 239 & 261 \\
\bottomrule
\end{tabular}
\caption{Realized chart structures in the three evaluation pools.}
\label{tab:app_structure}
\end{table*}

\begin{table*}[t]
\centering
\small
\setlength{\tabcolsep}{4.5pt}
\begin{tabular}{@{}lrrrrrrr@{}}
\toprule
& \multicolumn{3}{c}{Target category position}
& \multicolumn{4}{c}{Target series position} \\
\cmidrule(lr){2-4}\cmidrule(lr){5-8}
Pool & Left/first & Middle & Right/last
& First & Middle & Last & Single \\
\midrule
Axis scale    & 177 & 168 & 155 & 197 & 99 & 190 & 14 \\
Visual bundle & 149 & 182 & 169 & 229 & 89 & 182 & 0 \\
Prompt bundle & 167 & 165 & 168 & 190 & 98 & 212 & 0 \\
\bottomrule
\end{tabular}
\caption{
Relative positions of the queried category and series. Category positions
are grouped by normalized horizontal location: below one third, from one
third through two thirds, and above two thirds. Two-series charts contribute
first and last series positions; three-series charts additionally contribute
a middle position.
}
\label{tab:app_target_positions}
\end{table*}

For completeness, the target-series counts can be separated by chart
structure. In the axis pool, the 14 one-series cases all use index 0; the 184
two-series cases split 99/85 between indices 0/1; and the 302 three-series
cases split 98/99/105. The corresponding two-/three-series splits are
145/106 and 84/89/76 for the visual bundle, and 120/119 and 70/98/93 for the
prompt bundle. This confirms that the smaller aggregate count at series index
2 follows structural availability rather than omission of the final series
when it exists.

\subsubsection{Counterfactual directions and distances}

Edit directions are close to balanced rather than forced to be exactly equal.
Bar height always changes by one tick interval. Legend and series distances
are measured in series-index steps; x-label and category distances are
measured in category-index steps. Results are shown in Table~\ref{tab:app_intervention_directions}

\begin{table*}[t]
\centering
\small
\setlength{\tabcolsep}{4.2pt}
\renewcommand{\arraystretch}{1.10}
\begin{tabularx}{\textwidth}{@{}lX@{}}
\toprule
Intervention & Realized distribution over 500 queries \\
\midrule
Axis scale
& Pair orientation original-to-scaled/scaled-to-original: 263/237.
Directed tick-step transitions \(5{\rightarrow}10\),
\(10{\rightarrow}5\), \(10{\rightarrow}20\), and
\(20{\rightarrow}10\): 169, 157, 94, and 80. The scale ratio is two for
every pair. \\

Bar height
& Increase/decrease: 268/232. Signed changes \(+5,+10,-5,-10\):
135, 133, 121, and 111. Every edit changes the target by one tick interval. \\

Legend swap
& Higher/lower series index: 274/226. Index distance 1/2: 422/78. \\

X-label swap
& Left/right: 243/257. Index distances 1--5: 187, 165, 102, 39, and 7. \\

Prompt label
& Left/right: 259/241. Index distances 1--5: 233, 134, 85, 31, and 17. \\

Prompt series
& Higher/lower series index: 244/256. Index distance 1/2: 421/79. \\

Prompt label + series
& Label left/right: 248/252; series higher/lower: 237/263.
Label distances 1--5: 201, 155, 93, 39, and 12. Series distance 1/2:
413/87. \\
\bottomrule
\end{tabularx}
\caption{Realized counterfactual directions and index distances.}
\label{tab:app_intervention_directions}
\end{table*}

\subsubsection{ROI area audit}

Table~\ref{tab:app_roi_areas} reports ROI area as a fraction of the rendered
image. This normalized quantity is comparable across input resolutions,
unlike raw pixel counts. The bar-difference ROI is the smallest image region
in both models, covering approximately 0.6\% of the image on average.

\begin{table*}[t]
\centering
\small
\setlength{\tabcolsep}{4.8pt}
\begin{tabular}{@{}lrrrrrrrr@{}}
\toprule
& \multicolumn{4}{c}{\intern{} 448}
& \multicolumn{4}{c}{\qwen{} 1344} \\
\cmidrule(lr){2-5}\cmidrule(lr){6-9}
ROI & Mean & Median & Min & Max & Mean & Median & Min & Max \\
\midrule
Axis scale       & 3.048 & 2.759 & 2.351 & 4.314 & 2.309 & 2.089 & 1.784 & 3.274 \\
Bar union        & 2.462 & 2.337 & 0.711 & 5.273 & 2.106 & 1.951 & 0.565 & 4.677 \\
Bar difference   & 0.667 & 0.648 & 0.293 & 1.130 & 0.581 & 0.563 & 0.245 & 1.015 \\
Bar intersection & 1.795 & 1.752 & 0.303 & 4.717 & 1.524 & 1.475 & 0.219 & 4.172 \\
Legend swap      & 0.918 & 0.930 & 0.685 & 1.085 & 0.684 & 0.726 & 0.531 & 0.776 \\
X-label swap     & 1.336 & 1.334 & 1.321 & 1.347 & 1.122 & 1.124 & 1.120 & 1.124 \\
\bottomrule
\end{tabular}
\caption{ROI area as a percentage of the rendered image.}
\label{tab:app_roi_areas}
\end{table*}

\paragraph{Audit summary.}
The synchronized evaluation set covers the full realized target-value range,
all chart structures permitted by each intervention pool, all target-position
groups, both edit directions, and multiple edit distances. The principal
within-bundle results use paired queries, while source-chart cluster and
chart-equal analyses account for repeated source charts. These checks support
the reported conclusions for jointly successful original--counterfactual
cases; failure-case mechanisms are outside the scope of this evaluation.

\subsection{Factorial Recombination Experiment data generation and auditing}
\label{sec:fredaa}
\subsubsection{Initial Family Generation}

We first generated 10,000 families and rendered each family at two resolutions: $1344\times1344$ for Qwen2.5-VL-7B and $448\times448$ for InternVL3.5-8B. Apart from input resolution, the two model-specific versions use exactly the same chart values, series names, category names, colors, layout, prompt, target position, and receiver--donor assignment. Each family contains four charts that differ only in target-bar geometry and axis scale. Every chart contains two series and five categories, and the queried category is sampled from the three middle positions.

Let the base tick interval be $s\in\{5,10\}$ and the initial target value be $v$. Crossing two target-bar heights with two axis-scale settings gives the four charts in Table~\ref{tab:factorial-cells}.

Changing bar geometry moves only the top of the target bar by one tick interval; all other bars and the axis scale remain unchanged. Changing axis scale multiplies all bar values, the y-axis tick interval, and the y-axis upper limit by two. This preserves the rendered geometry of every bar, so the visible change is concentrated in the y-axis labels and their associated scale support.

Receiver assignments are balanced across the four chart types and shuffled with a fixed random seed. Relative to the assigned receiver, the geometry donor switches only the geometry state, the scale donor switches only the scale state, and the diagonal switches both. Because every chart type can serve as the receiver, the evaluation includes both increases and decreases in bar height and axis scale.

The generation pipeline also checks that the four correct answers are distinct. It also confirms that the target-bar and y-axis scale regions do not overlap at either the pixel level or the model's merged image-token grids.

\subsubsection{Model Inference and Final Family Selection}

We next asked Qwen2.5-VL-7B and InternVL3.5-8B to answer all four unpatched charts in every family. A response was counted as correct only if it could be parsed as a single integer and exactly matched the chart's correct answer.

Of the 10,000 generated families, Qwen2.5-VL-7B answered all four charts correctly in 5,963 families, and InternVL3.5-8B did so in 7,264 families. Both models answered all four charts correctly in 4,903 families. We selected 500 of these cross-model eligible families for activation patching, balancing both receiver type and base tick interval. Selection used only unpatched answer correctness and did not use any activation-patching result.

\subsubsection{Evaluation Data Distribution}

\begin{table}[t]
\centering
\small
\begin{tabularx}{\columnwidth}{@{}lX@{}}
\toprule
Data dimension & Distribution in the final 500 families \\
\midrule
Receiver type & $C_{00}$, $C_{10}$, $C_{01}$, and $C_{11}$: 125 each \\
Base tick interval & 5 and 10: 250 each \\
Bar-height direction & Increase: 250; decrease: 250 \\
Axis-scale direction & Increase: 250; decrease: 250 \\
Target series position & First series: 246; second series: 254 \\
Target category position & Second category: 150; third: 153; fourth: 197 \\
Four candidate values & Pooled range: 11--90; median: 32.0 \\
Within-family candidate span & Range: 21--55; mean: 35.752 \\
\bottomrule
\end{tabularx}
\caption{Distribution of the 500 families used in the factorial recombination
experiments. Target categories were sampled only from the three middle
positions of each five-category chart. ``All candidate values (pooled)''
combines the four candidate values from all families, giving 2,000 values
in total. ``Within-family candidate span'' is the difference between the
largest and smallest candidate values in each family; its minimum,
maximum, and mean summarize the 500 resulting spans.}
\label{tab:factorial-data-distribution}
\end{table}

The final evaluation set distribution is shown in Table~\ref{tab:factorial-data-distribution}. The final evaluation set is balanced across receiver types, base tick intervals, and the directions of both factor changes. The two target-series positions are nearly evenly represented, all three eligible target-category positions are covered, and the pooled candidate range of 11--90. The final data therefore do not collapse onto one receiver type, one change direction, one query position, or a narrow value range.

This evaluation set is nevertheless success-conditioned: every selected family had to be answered correctly by both models before intervention. The conclusions therefore apply to this clean-correct distribution and should not be directly extended to families on which either model fails without intervention.

\clearpage

\subsection{Formal Definitions and Computation of Statistical Metrics}
\label{app:metric_definitions}

This section formalizes the quantities used in the main analysis and
states the order in which they are computed. Unless otherwise noted, a
statistic is first computed for each query and is then aggregated across
queries. Let $i\in\{1,\ldots,N\}$ index queries, let
$k\in\{1,\ldots,K\}$ index intervention conditions, and let $g(i)$ be
the source chart associated with query $i$. A source chart can
contribute several series--category queries.

\subsubsection{Answer-sequence score and answer margin}

\paragraph{Definition.}
Let $a_i^{o}$ and $a_i^{c}$ be the original and counterfactual gold
answer sequences. For a run $x\in\{o,c,p\}$, denoting the original,
counterfactual, and patched computations, respectively, the
teacher-forced score of an answer sequence $a$ is
\begin{equation}
S_i^{x}(a)
=
\sum_{t=1}^{T(a)}
\log P_{\theta}
\bigl(a_t\mid x_i,q_i,a_{<t}\bigr).
\label{eq:app_sequence_score}
\end{equation}
The answer-preference margin in run $x$ is
\begin{equation}
M_i^{x}
=
S_i^{x}(a_i^{o})-S_i^{x}(a_i^{c}).
\label{eq:app_answer_margin}
\end{equation}

\paragraph{Explanation.}
An integer answer can contain more than one output token. Scoring the
complete answer sequence avoids reducing a multi-token number to its
first token. The margin compares the only two answers relevant to the
paired intervention: the original gold answer and the counterfactual
gold answer. A positive margin favors the original answer, while a
negative margin favors the counterfactual answer.

\subsubsection{Normalized restoration}

\paragraph{Definition.}
For query $i$, normalized restoration is
\begin{equation}
R_i
=
\frac{M_i^{p}-M_i^{c}}
     {M_i^{o}-M_i^{c}}.
\label{eq:app_normalized_restoration}
\end{equation}

Restoration values are not clipped to $[0,1]$.

\paragraph{Explanation.}
The denominator is the full answer-preference change between the
counterfactual and original runs. The numerator is the part of that
change recovered by patching. Thus, $R_i=0$ means that patching leaves
the counterfactual preference unchanged, while $R_i=1$ means that the
full original--counterfactual margin gap is restored. A value above one
indicates overshooting, and a negative value indicates movement further
toward the counterfactual answer. Normalization makes effects more
comparable across queries and models with different raw log-probability
scales.

\subsubsection{Projected-site, layer-wise, and prompt restoration}

\paragraph{Definition.}
For image intervention $k$, let $\Omega_{i,k}$ be its pixel-level region
of interest (ROI), and let
\begin{equation}
\mathcal{J}_{i,k}^{m}
=
\mathcal{T}_{m}(\Omega_{i,k})
\label{eq:app_roi_token_mapping}
\end{equation}
be the visual-token set selected by spatial mapping rule $m$.
Projected-site restoration is the normalized restoration obtained by
replacing the selected counterfactual projected tokens with their
original counterparts:
\begin{equation}
\begin{aligned}
R_{i,k}^{\mathrm{proj},m}
=
R_i\bigl(&h_{i,\mathrm{proj}}^{c}
          [\mathcal{J}_{i,k}^{m}]\\[-1mm]
        &\leftarrow
          h_{i,\mathrm{proj}}^{o}
          [\mathcal{J}_{i,k}^{m}]\bigr).
\end{aligned}
\label{eq:app_projected_restoration}
\end{equation}
At language layer $\ell$, image-token restoration is denoted
\begin{equation}
\begin{aligned}
R_{i,k,\ell}^{m}
=
R_i\bigl(&h_{i,\ell}^{c}
          [\mathcal{J}_{i,k}^{m}]\\[-1mm]
        &\leftarrow
          h_{i,\ell}^{o}
          [\mathcal{J}_{i,k}^{m}]\bigr).
\end{aligned}
\label{eq:app_layer_restoration}
\end{equation}
For prompt intervention $k$, let $\mathcal{P}_{i,k}$ be the changed
prompt-token positions. Prompt restoration at layer $\ell$ is
\begin{equation}
\begin{aligned}
R_{i,k,\ell}^{\mathrm{prompt}}
=
R_i\bigl(&h_{i,\ell}^{c}[\mathcal{P}_{i,k}]\\[-1mm]
        &\leftarrow
          h_{i,\ell}^{o}[\mathcal{P}_{i,k}]\bigr).
\end{aligned}
\label{eq:app_prompt_restoration}
\end{equation}

\paragraph{Explanation.}
Projected-site restoration asks how much of the original answer
preference can be recovered using only the visual tokens associated
with one semantic region, such as the changed bar top, invariant bar
body, legend entries, x-axis labels, or y-axis scale. Layer-wise
restoration repeats the same intervention at different language layers
to measure where the local state remains causally recoverable. Prompt
interventions patch category or series token positions while keeping
the image fixed; they therefore do not have a projected-image condition.

\subsubsection{Query-weighted mean}

\paragraph{Definition.}
For a query-level statistic $Z_i$, the query-weighted estimate is
\begin{equation}
\widehat{\mu}_{\mathrm{query}}
=
\frac{1}{N}\sum_{i=1}^{N}Z_i.
\label{eq:app_query_mean}
\end{equation}

\paragraph{Explanation.}
This estimate gives every series--category question equal weight. It
answers: if one query is drawn from the evaluation pool, what effect is
expected on average? It is the primary estimand because each patching
record corresponds to one concrete value-reading query.

\subsubsection{Query-level bootstrap confidence interval}

\paragraph{Definition.}
For bootstrap replicate $b\in\{1,\ldots,B\}$, sample $N$ query indices
with replacement and compute
\begin{equation}
\widehat{\mu}^{*(b)}
=
\frac{1}{N}
\sum_{j=1}^{N}Z_{i_{b,j}^{*}}.
\label{eq:app_bootstrap_replicate}
\end{equation}
We use $B=5{,}000$ replicates with random seed 42. Define the bootstrap
sample compactly as
\begin{equation}
\mathcal{B}_{\mu}
=
\{\widehat{\mu}^{*(b)}:b=1,\ldots,B\}.
\label{eq:app_bootstrap_sample}
\end{equation}
The percentile interval is
\begin{equation}
\begin{aligned}
\mathrm{CI}_{95\%}
=
\bigl[
 Q_{0.025}(\mathcal{B}_{\mu}),
 Q_{0.975}(\mathcal{B}_{\mu})
\bigr],
\end{aligned}
\label{eq:app_bootstrap_ci}
\end{equation}
where $Q_{\alpha}$ is the empirical $\alpha$-quantile.

\paragraph{Explanation.}
The bootstrap estimates how much the reported mean would vary if a
similar set of queries were sampled again. It does not require the
restoration distribution to be Gaussian, which is useful because
normalized restoration can be negative or greater than one. For an AUC
comparison, the complete AUC is first computed for every query; the
resulting query-level AUC values, rather than individual layer points,
are then resampled.

\subsubsection{Paired condition and ROI contrasts}

\paragraph{Definition.}
Suppose conditions $A$ and $B$ use the same query keys. Their
within-query difference is
\begin{equation}
\Delta_i^{A-B}
=
Z_i^{A}-Z_i^{B}.
\label{eq:app_paired_difference}
\end{equation}
The paired mean difference is
\begin{equation}
\widehat{\Delta}^{A-B}
=
\frac{1}{N}
\sum_{i=1}^{N}\Delta_i^{A-B}.
\label{eq:app_paired_mean}
\end{equation}
The positive-direction fraction is
\begin{equation}
\widehat{\pi}_{+}^{A-B}
=
\frac{1}{N}
\sum_{i=1}^{N}
\mathbf{1}[\Delta_i^{A-B}>0].
\label{eq:app_positive_fraction}
\end{equation}
Confidence intervals are formed by bootstrapping the paired differences
$\{\Delta_i^{A-B}\}_{i=1}^{N}$ rather than the two conditions
independently.

\paragraph{Explanation.}
Pairing removes variation caused by different charts, targets, or
counterfactual edits. Bar difference and bar intersection, for example,
use the same original--counterfactual image pair and differ only in the
patched ROI. Their paired contrast therefore asks whether the changed
bar top restores more information than the invariant body on the same
query. The paired mean measures the average size of this advantage,
while the positive-direction fraction measures how consistently the
same direction holds across individual queries.

\subsubsection{Source-chart cluster bootstrap}

\paragraph{Definition.}
Let
\begin{equation}
\mathcal{I}_{c}
=
\{i:g(i)=c\},
\qquad
n_c=\lvert\mathcal{I}_{c}\rvert
\label{eq:app_chart_cluster}
\end{equation}
be the queries derived from source chart $c$. In replicate $b$, sample
$C$ source-chart identifiers with replacement and include all queries
belonging to each sampled chart. The query-weighted replicate is
\begin{equation}
\begin{aligned}
\widehat{\mu}_{\mathrm{cl}}^{*(b)}
=
\frac{
 \sum_{j=1}^{C}
 \sum_{i\in\mathcal{I}_{c_{b,j}^{*}}} Z_i
}{
 \sum_{j=1}^{C}n_{c_{b,j}^{*}}
}.
\end{aligned}
\label{eq:app_cluster_bootstrap}
\end{equation}
The cluster-bootstrap interval is obtained from the 2.5th and 97.5th
percentiles of $B=5{,}000$ such replicates.

\paragraph{Explanation.}
Queries from the same source chart share layout, colors, labels, axis
configuration, and values. Treating them as fully independent could
make an uncertainty interval too narrow. Cluster bootstrap resamples
whole charts and keeps all associated queries together. It preserves
the query-weighted target of estimation while accounting for dependence
among queries from the same chart. For a paired comparison, $Z_i$ is
replaced with $\Delta_i^{A-B}$.

\subsubsection{Chart-equal estimate}

\paragraph{Definition.}
First compute the mean within source chart $c$:
\begin{equation}
\overline{Z}_c
=
\frac{1}{n_c}
\sum_{i\in\mathcal{I}_{c}}Z_i.
\label{eq:app_within_chart_mean}
\end{equation}
The chart-equal estimate is
\begin{equation}
\widehat{\mu}_{\mathrm{chart}}
=
\frac{1}{C}
\sum_{c=1}^{C}\overline{Z}_c.
\label{eq:app_chart_equal}
\end{equation}

\paragraph{Explanation.}
The query-weighted mean gives more influence to charts that contribute
more target bars. The chart-equal estimate instead gives every source
chart the same total weight. Agreement between the two estimates shows
that a result is not driven only by a small number of charts that happen
to generate many queries.

\subsubsection{ROI area and selected-token controls}

\paragraph{Definition.}
For ROI $r$, normalized pixel area is
\begin{equation}
A_{i,r}
=
\frac{1}{HW}
\sum_{u=1}^{H}\sum_{v=1}^{W}
\mathbf{1}[(u,v)\in\Omega_{i,r}].
\label{eq:app_roi_area}
\end{equation}
Under mapping rule $m$, the number of selected visual tokens is
\begin{equation}
N_{i,r}^{m}
=
\lvert\mathcal{T}_{m}(\Omega_{i,r})\rvert.
\label{eq:app_token_count}
\end{equation}
For the main changed-top comparison, define
\begin{equation}
\mathcal{S}_{\mathrm{tok}}^{m}
=
\{i:N_{i,\mathrm{diff}}^{m}
\leq N_{i,\mathrm{inter}}^{m}\}.
\label{eq:app_token_subset}
\end{equation}
The restricted paired effect is
\begin{equation}
\begin{aligned}
\widehat{\Delta}_{\mathrm{tok}}^{m}
=
\frac{1}{\lvert\mathcal{S}_{\mathrm{tok}}^{m}\rvert}
\sum_{i\in\mathcal{S}_{\mathrm{tok}}^{m}}
\bigl(
 R_{i,\mathrm{diff}}^{m}
 -R_{i,\mathrm{inter}}^{m}
\bigr).
\end{aligned}
\label{eq:app_token_control}
\end{equation}

\paragraph{Explanation.}
Patching more tokens can mechanically increase restoration because more
of the original state is inserted. The restricted comparison keeps only
queries for which the changed bar-top ROI selects no more tokens than
the invariant body. If the changed-top advantage remains, it cannot be
explained simply by patching a larger number of tokens. Pixel area and
token count are both useful because a small pixel region can intersect
several token cells.

\subsubsection{Fraction of the whole-bar effect}

\paragraph{Definition.}
For queries with
$\lvert R_{i,\mathrm{union}}\rvert\geq\epsilon$, define
\begin{align}
Q_{i,\mathrm{diff}}
&=
\frac{R_{i,\mathrm{diff}}}
     {R_{i,\mathrm{union}}},
\label{eq:app_diff_union_ratio}\\
Q_{i,\mathrm{inter}}
&=
\frac{R_{i,\mathrm{inter}}}
     {R_{i,\mathrm{union}}}.
\label{eq:app_inter_union_ratio}
\end{align}
The analysis reports the sample mean and median of these per-query
ratios.

\paragraph{Explanation.}
The union ROI covers the full affected bar. These ratios ask how much of
that whole-bar effect is already captured by the changed top or by the
invariant body. A changed-top-to-union ratio near one means that the
narrow changed band restores almost as much answer preference as the
whole-bar patch. Reporting both mean and median makes the summary less
dependent on a few unusually large ratios.

\subsubsection{Normalized layer depth}

\paragraph{Definition.}
For a model with $L$ language layers indexed by
$\ell\in\{0,\ldots,L-1\}$, normalized depth is
\begin{equation}
d_{\ell}
=
\frac{\ell}{L-1}.
\label{eq:app_normalized_depth}
\end{equation}

\paragraph{Explanation.}
The evaluated models have different layer counts, so raw layer indices
are not directly comparable. Normalized depth maps the first layer to
zero and the final layer to one. It allows transition locations to be
compared as fractions of total network depth, but it does not imply that
two layers at the same normalized depth perform identical computations.

\subsubsection{Raw layer-profile AUC}

\paragraph{Definition.}
Let $R_{i,k,\ell}$ be single-layer restoration. The per-query raw AUC
is computed using the trapezoidal rule:
\begin{equation}
\begin{aligned}
\mathrm{AUC}_{i,k}^{\mathrm{raw}}
=
\sum_{\ell=0}^{L-2}
&\frac{R_{i,k,\ell}+R_{i,k,\ell+1}}{2}\\[-1mm]
&\times(d_{\ell+1}-d_{\ell}).
\end{aligned}
\label{eq:app_raw_auc}
\end{equation}
A paired AUC advantage of condition $A$ over condition $B$ is
\begin{equation}
\Delta_{i,\mathrm{AUC}}^{A-B}
=
\mathrm{AUC}_{i,A}^{\mathrm{raw}}
-
\mathrm{AUC}_{i,B}^{\mathrm{raw}}.
\label{eq:app_raw_auc_difference}
\end{equation}

\paragraph{Explanation.}
Raw AUC summarizes the total restoration accumulated over the full
network depth. It is large when an ROI starts with a strong restoration
effect, when that effect remains accessible across many layers, or
both. It therefore measures cumulative recoverable information, but it
does not separate initial magnitude from persistence. AUC is computed
for each query before averaging, pairing, or bootstrapping; this is the
quantity referred to as the raw layer-profile AUC in the prompt-label
versus prompt-series comparison.

\subsubsection{Retention profile and retention AUC}

\paragraph{Definition.}
For queries whose layer-0 restoration is nonzero at numerical precision,
define
\begin{equation}
\mathcal{E}_k
=
\{i:\lvert R_{i,k,0}\rvert\geq\epsilon\}.
\label{eq:app_retention_eligible}
\end{equation}
For each eligible query, the retention profile is
\begin{equation}
\rho_{i,k,\ell}
=
\frac{R_{i,k,\ell}}{R_{i,k,0}}.
\label{eq:app_retention_profile}
\end{equation}
Its per-query retention AUC is
\begin{equation}
\begin{aligned}
\mathrm{AUC}_{i,k}^{\mathrm{ret}}
=
\sum_{\ell=0}^{L-2}
&\frac{\rho_{i,k,\ell}+\rho_{i,k,\ell+1}}{2}\\[-1mm]
&\times(d_{\ell+1}-d_{\ell}).
\end{aligned}
\label{eq:app_retention_auc}
\end{equation}

\paragraph{Explanation.}
Two ROIs can begin with very different restoration magnitudes. Dividing
each query's curve by its own layer-0 value sets every eligible curve to
one at layer zero, so the remaining quantity measures the fraction of
the initial local effect that remains recoverable. Raw AUC asks how much
recoverable information accumulates across depth; retention AUC asks how
slowly the initially available effect decays. These two metrics must be
reported together because a weak effect can persist for a long time and
a strong early effect can disappear quickly.

\subsubsection{Relative half-life}

\paragraph{Definition.}
The mean retention curve for condition $k$ is
\begin{equation}
\overline{\rho}_{k,\ell}
=
\frac{1}{\lvert\mathcal{E}_k\rvert}
\sum_{i\in\mathcal{E}_k}
\rho_{i,k,\ell}.
\label{eq:app_mean_retention}
\end{equation}
The half-life layer is the first native layer position at which this
mean curve reaches or falls below one half:
\begin{equation}
\ell_{1/2}^{k}
=
\min\{\ell:\overline{\rho}_{k,\ell}\leq0.5\}.
\label{eq:app_half_life_layer}
\end{equation}
Relative half-life is
\begin{equation}
H_k
=
d_{\ell_{1/2}^{k}}
=
\frac{\ell_{1/2}^{k}}{L-1}.
\label{eq:app_relative_half_life}
\end{equation}
If the mean curve never reaches $0.5$, half-life is reported as not
reached.

\paragraph{Explanation.}
Relative half-life provides one readable location for the decay of a
retention curve: it is the relative network depth at which less than
half of the initial local effect remains. The reported values align with
native normalized layer positions; for example, $22/27\approx0.815$ for
a 28-layer model and $23/35\approx0.657$ for a 36-layer model. Half-life
is only a summary of the curve. It is not the exact layer at which the
model performs a computation, because information can be transformed or
redistributed and thereby become inaccessible to a particular local
patch.

\subsubsection{Contiguous layer-window patching}

\paragraph{Definition.}
For width $w\in\{2,4\}$ and starting layer
$s\in\{0,\ldots,L-w\}$, define
\begin{equation}
\mathcal{W}_{s,w}
=
\{s,s+1,\ldots,s+w-1\}.
\label{eq:app_window_set}
\end{equation}
All selected token states in this window are replaced jointly in one
forward pass. We denote the resulting normalized restoration by
\begin{equation}
R_{i,k,s}^{(w)}
=
R_i\bigl(
\operatorname{Patch}
(\mathcal{J}_{i,k},\mathcal{W}_{s,w})
\bigr).
\label{eq:app_window_restoration}
\end{equation}
Windows use stride one. Their normalized center depth is
\begin{equation}
d_{s,w}^{\mathrm{center}}
=
\frac{s+(w-1)/2}{L-1}.
\label{eq:app_window_center}
\end{equation}

\paragraph{Explanation.}
A representation can be distributed across neighboring layers, so a
single-layer intervention may underestimate how long it remains
recoverable. Window patching jointly restores a short contiguous range
to test whether the ordering and transition regions remain visible.
The result is produced by a new multi-layer forward intervention; it is
not the sum or average of the corresponding single-layer effects. A
larger window effect is therefore interpreted as accumulated recovery,
not as evidence that every layer in the window performs the same
operation.

\subsubsection{Spatial-expansion Gain and Within-Model Profile Agreement}
\label{app:spatial_mapping_agreement}

This analysis compares the two spatial ROI-to-token mappings evaluated
for \qwen{}: strict any-overlap mapping and merge-closed
$3{\times}3$ coarse mapping. The comparison contains two distinct
quantities. Projected-site gain measures a change in restoration
magnitude at the projected image-token site, whereas within-model
profile correlation measures whether the two mappings produce similar
restoration trajectories across the model's native language layers.

\paragraph{Definition: projected-site restoration gain.}
Let
\[
m\in\{\mathrm{any},\mathrm{coarse}\}
\]
denote the spatial mapping rule. For an image intervention $k$, query
$i$, and mapping $m$, projected-site restoration
$R_{i,k}^{\mathrm{proj},m}$ is defined in
Equation~\ref{eq:app_projected_restoration}. The per-query gain from
coarse mapping is
\begin{equation}
G_{i,k}^{\mathrm{proj}}
=
R_{i,k}^{\mathrm{proj},\mathrm{coarse}}
-
R_{i,k}^{\mathrm{proj},\mathrm{any}}.
\label{eq:app_coarse_projected_gain}
\end{equation}
Its paired mean is
\begin{equation}
\widehat{G}_{k}^{\mathrm{proj}}
=
\frac{1}{N}
\sum_{i=1}^{N}
G_{i,k}^{\mathrm{proj}}.
\label{eq:app_mean_coarse_projected_gain}
\end{equation}

\paragraph{Explanation.}
This quantity compares the two mappings at one fixed patching site:
the projected image-token representation before subsequent
language-layer processing. A positive value means that replacing the
coarse ROI-selected projected tokens restores more of the original
answer preference than replacing the tokens selected by strict
any-overlap mapping. It measures a difference in restoration
magnitude, not a difference in the shape or timing of a layer-wise
profile. Prompt interventions have no projected-site gain because
their patched supports are prompt-token positions rather than image
ROIs.

\paragraph{Definition: mapping-specific mean layer profiles.}
For an image intervention $k$, the layer-wise restoration under mapping
$m$ is $R_{i,k,\ell}^{m}$, as defined in
Equation~\ref{eq:app_layer_restoration}. At native language layer
$\ell$, its query-averaged profile value is
\begin{equation}
\overline{R}_{k,\ell}^{m}
=
\frac{1}{N}
\sum_{i=1}^{N}
R_{i,k,\ell}^{m}.
\label{eq:app_mapping_mean_profile}
\end{equation}
The complete mean profile is
\begin{equation}
\overline{\mathbf{R}}_{k}^{m}
=
\left(
\overline{R}_{k,0}^{m},
\ldots,
\overline{R}_{k,L-1}^{m}
\right).
\label{eq:app_mapping_profile_vector}
\end{equation}

\paragraph{Explanation.}
Each element of
$\overline{\mathbf{R}}_{k}^{m}$ is obtained by patching the
ROI-selected image-token positions at one particular language layer
and then averaging restoration over the synchronized queries. The
vector therefore describes how the locally patched information remains
recoverable across language-model depth. It is separate from
projected-site restoration, which produces one value per query at the
fixed projected image-token site.

\paragraph{Definition: within-model profile correlation.}
Because both mappings are evaluated in the same \qwen{} model, they
share the same $L$ native language layers. Their profile-shape agreement
for image intervention $k$ is
\begin{equation}
r_{k}^{\mathrm{map}}
=
\operatorname{corr}_{\mathrm{P}}
\left(
\overline{\mathbf{R}}_{k}^{\mathrm{any}},
\overline{\mathbf{R}}_{k}^{\mathrm{coarse}}
\right),
\label{eq:app_mapping_profile_corr}
\end{equation}
where $\operatorname{corr}_{\mathrm{P}}$ denotes Pearson correlation
across the $L$ native layer positions. No normalized-depth interpolation
is required for this within-model comparison.

\paragraph{Explanation.}
The correlation measures whether the two average layer-wise curves rise,
decline, and approach zero at similar native layers. Pearson correlation
mainly captures profile shape rather than absolute magnitude. The coarse
profile can therefore have larger restoration values while still
showing a correlation close to one with the any-overlap profile.

A high value of $r_{k}^{\mathrm{map}}$ does not mean that the two
mappings patch the same tokens or produce equal restoration at every
layer. It means only that changing the ROI-to-token mapping leaves the
main depth-dependent trajectory largely unchanged.

\subsubsection{Cross-model profile and query-level correlations}
\label{app:cross_model_corr}

\paragraph{Definition: native mean profiles.}
For model $m$ with $L_m$ layers, the condition-level mean restoration
at native layer $\ell$ is
\begin{equation}
\overline{R}_{k,\ell}^{(m)}
=
\frac{1}{N}
\sum_{i=1}^{N}R_{i,k,\ell}^{(m)},
\label{eq:app_native_mean_profile}
\end{equation}
and its normalized depth is
\begin{equation}
d_{\ell}^{(m)}
=
\frac{\ell}{L_m-1}.
\label{eq:app_model_normalized_depth}
\end{equation}

\paragraph{Definition: common grid and interpolation.}
We use the common grid
\begin{equation}
\mathcal{D}_{G}
=
\{d_j=j/G:j=0,\ldots,G\},
\qquad G=100.
\label{eq:app_common_depth_grid}
\end{equation}
Thus, both mean profiles are represented at 101 equally spaced relative
depths. For a grid point $d_j$ lying between two adjacent native depths,
$d_{\ell}^{(m)}\leq d_j\leq d_{\ell+1}^{(m)}$, define
\begin{equation}
\lambda_{j,\ell}^{(m)}
=
\frac{d_j-d_{\ell}^{(m)}}
     {d_{\ell+1}^{(m)}-d_{\ell}^{(m)}}.
\label{eq:app_interp_weight}
\end{equation}
The piecewise-linear interpolated value is
\begin{equation}
\begin{aligned}
\widetilde{R}_{k}^{(m)}(d_j)
={}&
(1-\lambda_{j,\ell}^{(m)})
\overline{R}_{k,\ell}^{(m)}\\[-1mm]
&+
\lambda_{j,\ell}^{(m)}
\overline{R}_{k,\ell+1}^{(m)}.
\end{aligned}
\label{eq:app_linear_interpolation}
\end{equation}
At $d_0=0$ and $d_G=1$, the original first- and last-layer values are
used directly.

\paragraph{Definition: profile correlation.}
Let
\begin{equation}
\widetilde{\mathbf{R}}_{k}^{(m)}
=
\{\widetilde{R}_{k}^{(m)}(d_j)\}_{j=0}^{G}.
\label{eq:app_interpolated_profile_vector}
\end{equation}
For models $A$ and $B$, profile similarity is
\begin{equation}
r_{k}^{\mathrm{profile}}
=
\operatorname{corr}_{\mathrm{P}}
\bigl(
\widetilde{\mathbf{R}}_{k}^{(A)},
\widetilde{\mathbf{R}}_{k}^{(B)}
\bigr).
\label{eq:app_profile_correlation}
\end{equation}
The mean correlation across the $K$ intervention types is
\begin{equation}
\overline{r}^{\mathrm{profile}}
=
\frac{1}{K}
\sum_{k=1}^{K}r_{k}^{\mathrm{profile}}.
\label{eq:app_mean_profile_correlation}
\end{equation}

\paragraph{Explanation: what profile correlation measures.}
Each point entering $r_{k}^{\mathrm{profile}}$ is a relative-depth
position on an average curve, not an individual chart. A high value
means that the two mean curves tend to stay high, decline, and approach
zero at similar relative depths. Pearson correlation is mainly a
measure of curve shape: two curves can have different absolute
magnitudes and still be highly correlated if they rise and fall in
similar places. The interpolation only aligns the horizontal depth
axis. It does not create an actual intermediate layer or imply a
one-to-one functional correspondence between the two architectures.

\paragraph{Definition: per-query projected-restoration correlation.}
For synchronized queries, the cross-model sample correlation at the
projected image-token site is
\begin{equation}
\begin{aligned}
r_{k}^{\mathrm{query}}
=
\operatorname{corr}_{\mathrm{P}}
\bigl(&
\{R_{i,k}^{\mathrm{proj},A}\}_{i=1}^{N},\\[-1mm]
&
\{R_{i,k}^{\mathrm{proj},B}\}_{i=1}^{N}
\bigr).
\end{aligned}
\label{eq:app_query_correlation}
\end{equation}
No layer interpolation is used here: every query contributes one
projected-site restoration value from each model.

\paragraph{Explanation: why the two correlations can differ.}
The profile correlation first averages over all queries at each layer,
so chart-specific differences are largely smoothed out. It asks whether
the two models share the same broad population-level depth pattern. The
query-level correlation keeps the individual examples separate. It asks
whether a chart with unusually high restoration in one model also has
unusually high restoration in the other. The two models can therefore
have a high profile correlation but a low query-level correlation: they
can agree on the average ordering and timing of information loss while
disagreeing about which exact charts are most locally recoverable. Low
query-level agreement can reflect architectural differences, visual
token grids, spatial mappings, and model-specific answer-margin scales.
Neither correlation by itself establishes that the two models implement
an identical internal mechanism.

\clearpage

\begin{table*}[t]
\centering
\footnotesize
\setlength{\tabcolsep}{4.5pt}
\renewcommand{\arraystretch}{1.15}
\begin{tabularx}{\textwidth}{
    @{}
    >{\raggedright\arraybackslash}p{1.35cm}
    >{\raggedright\arraybackslash}p{1.65cm}
    >{\raggedright\arraybackslash}X
    >{\raggedright\arraybackslash}X
    @{}
}
\toprule
Modality
& Counterfactual
& Counterfactual generation
& Patched support / ROI definition \\
\midrule

\multirow{6}{*}{Image}
& Axis scale
& Multiply all y-axis tick values and the corresponding target value by \(2\), while preserving the rendered bar geometry. In the formal patching experiments, the original and counterfactual members are randomly exchanged for approximately half of the pairs.
& The y-axis scale region, including the affected tick labels and their associated scale support. Example is shown in Figure~\ref{fig:vertical-bar-axis-scale}. \\

\cmidrule(lr){2-4}

& \multirow{3}{*}{Bar height}
& \multirow{3}{=}{Randomly increase or decrease the target bar height by exactly one y-axis tick step, subject to the valid plotting range.}
& \textbf{Union:} union of the target-bar regions in the original and counterfactual images. Example is shown in Figure~\ref{fig:vertical-bar-bar-height-union} \\

&
&
& \textbf{Difference:} symmetric difference between the two target-bar regions, concentrated around the changed bar top. Example is shown in Figure~\ref{fig:vertical-bar-bar-height-difference} \\

&
&
& \textbf{Intersection:} intersection of the two target-bar regions, corresponding to the invariant shared bar body. Example is shown in Figure~\ref{fig:vertical-bar-bar-height-intersection} \\

\cmidrule(lr){2-4}

& Legend swap
& Randomly exchange the legend mapping of the target bar's series with that of another eligible series.
& The two affected legend-entry regions, including their visual keys and associated text labels. Example is shown in Figure~\ref{fig:vertical-bar-legend-swap}\\

\cmidrule(lr){2-4}

& X-label swap
& Randomly exchange the target bar's category label with another eligible category label.
& The two affected x-axis label slots. Example is shown in Figure~\ref{fig:vertical-bar-x-label-swap} \\

\midrule

\multirow{3}{*}{Prompt}
& Label
& Keep the image fixed and replace the target category expression with another valid category, while preserving token positions and token count required for activation patching.
& The prompt-token positions corresponding to the category expression. \\

\cmidrule(lr){2-4}

& Series
& Keep the image fixed and replace the target series expression with another valid series, with position-aligned prompt tokens.
& The prompt-token positions corresponding to the series expression. \\

\cmidrule(lr){2-4}

& Label + series
& Keep the image fixed and replace both the target category and target series expressions with position-aligned alternatives.
& The union of the prompt-token positions corresponding to the category and series expressions. \\

\bottomrule
\end{tabularx}
\caption{
Counterfactual construction and patched supports for the nine analysis
conditions. The three bar conditions share the same one-tick bar-height
counterfactual and differ only in the spatial ROI used for activation
patching. 
}
\label{tab:interventions}
\end{table*}

\begin{figure*}[t]
  \centering
  \includegraphics[width=\textwidth]{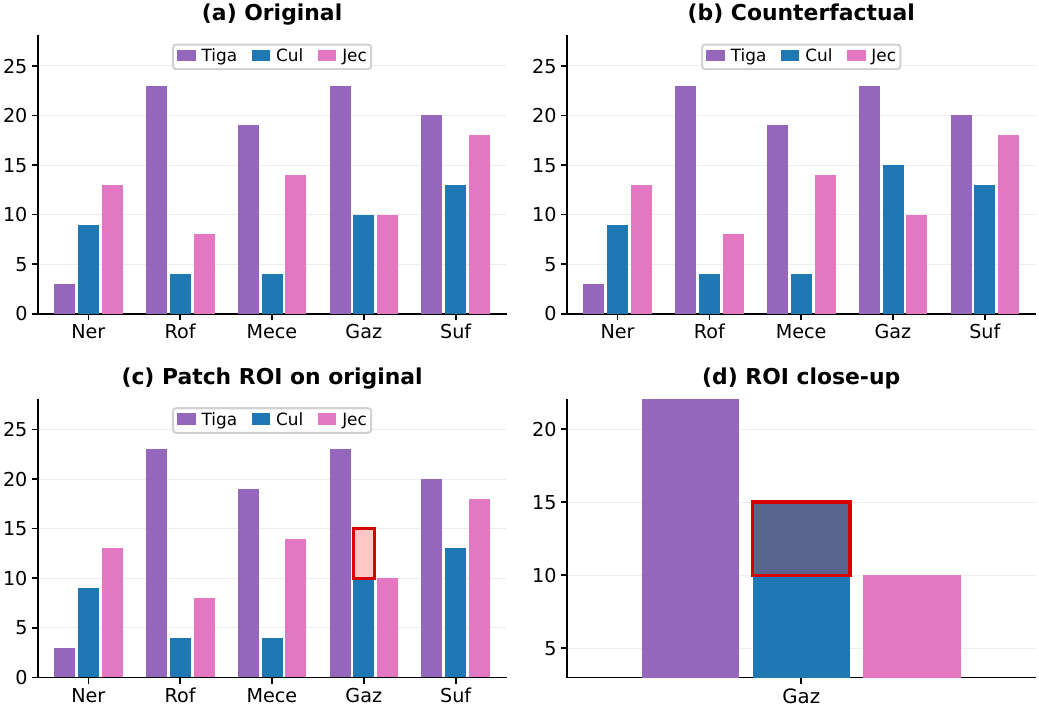}
  \caption{Example of the bar-height-difference counterfactual. The target bar value changes while the patch ROI covers only the changed height band.}
  \label{fig:vertical-bar-bar-height-difference}
\end{figure*}

\begin{figure*}[t]
  \centering
  \includegraphics[width=\textwidth]{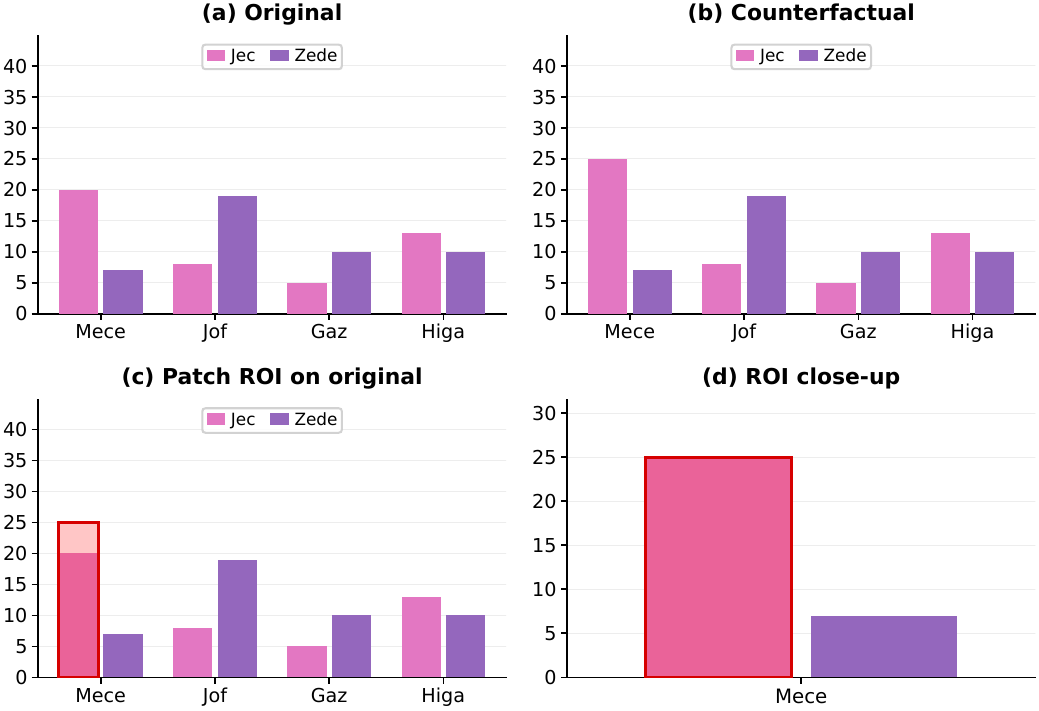}
  \caption{Example of the bar-height-union counterfactual. The target bar value changes while the patch ROI covers the geometric union of the original and counterfactual target bars.}
  \label{fig:vertical-bar-bar-height-union}
\end{figure*}

\begin{figure*}[t]
  \centering
  \includegraphics[width=\textwidth]{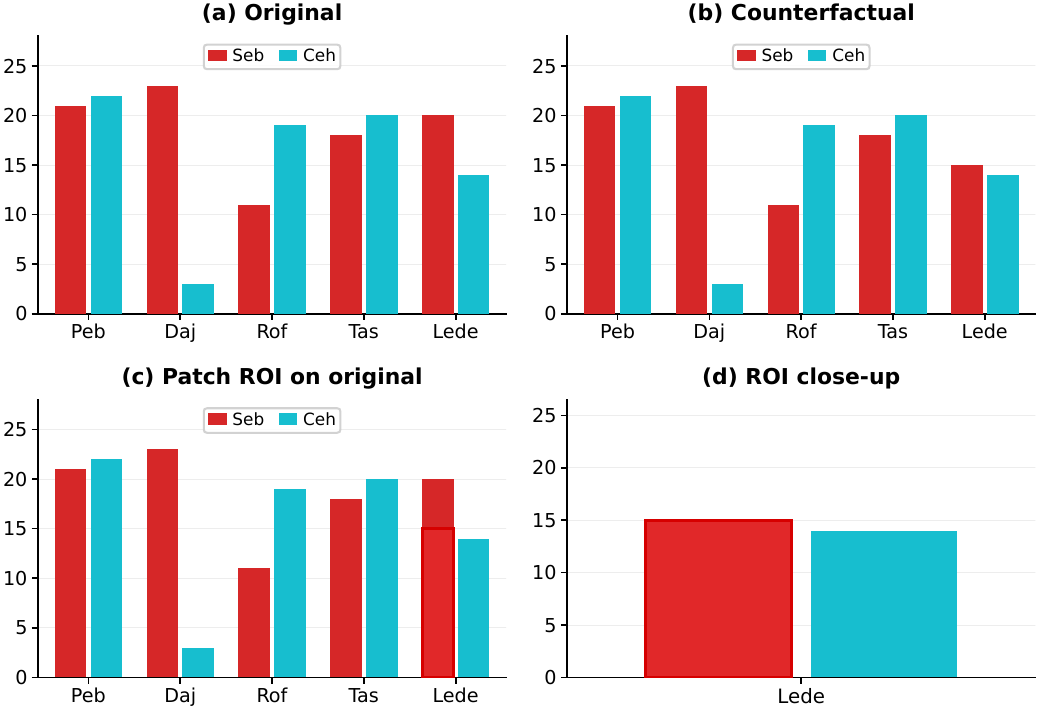}
  \caption{Example of the bar-height-intersection counterfactual. The target bar value changes while the patch ROI covers the geometric intersection of the original and counterfactual target bars.}
  \label{fig:vertical-bar-bar-height-intersection}
\end{figure*}

\begin{figure*}[t]
  \centering
  \includegraphics[width=\textwidth]{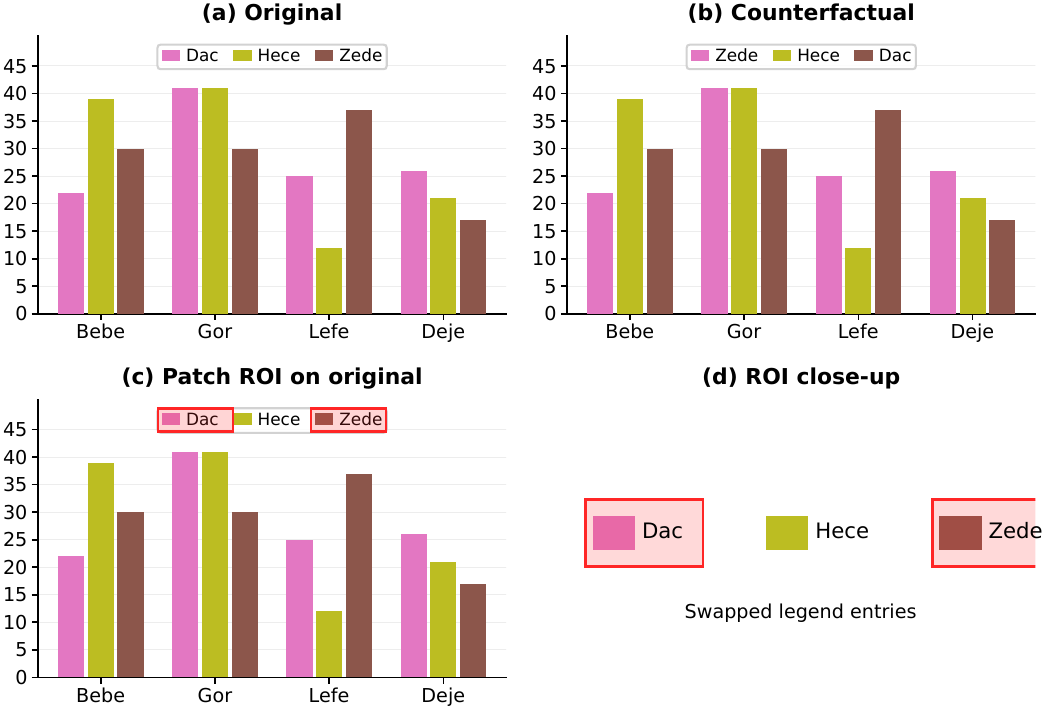}
  \caption{Example of the legend-swap counterfactual. The semantic perturbation is induced by swapping legend items; the patch ROI covers the swapped legend entries.}
  \label{fig:vertical-bar-legend-swap}
\end{figure*}

\begin{figure*}[t]
  \centering
  \includegraphics[width=\textwidth]{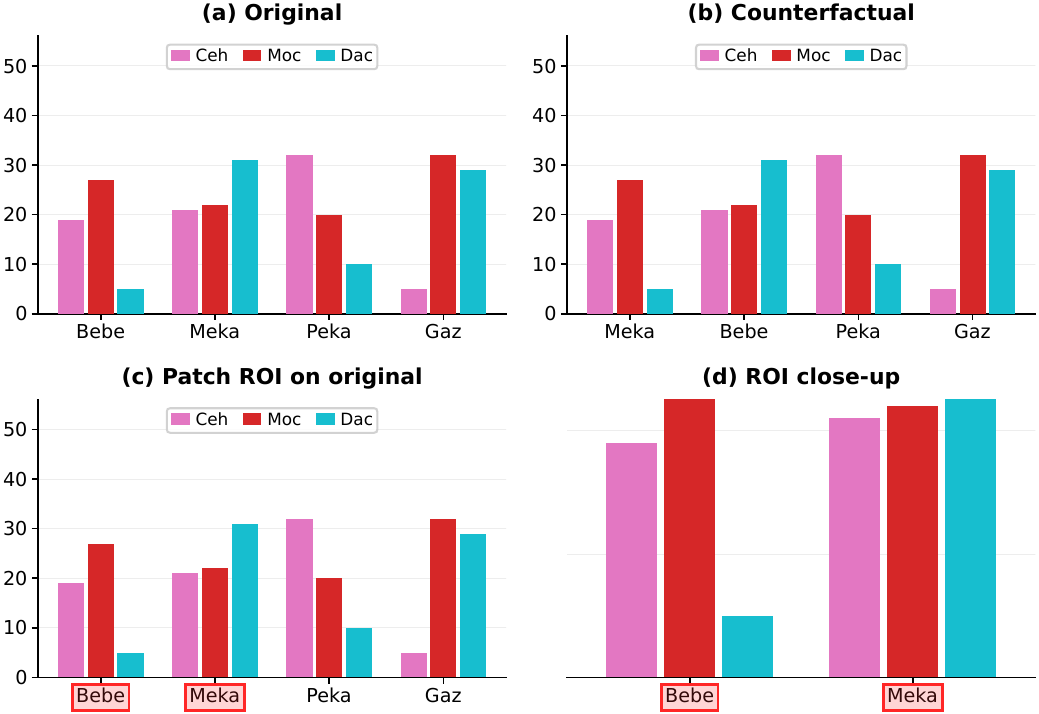}
  \caption{Example of the x-label-swap counterfactual. The semantic perturbation is induced by swapping two x-axis labels; the patch ROI covers the swapped label slots.}
  \label{fig:vertical-bar-x-label-swap}
\end{figure*}

\begin{figure*}[t]
  \centering
  \includegraphics[width=\textwidth]{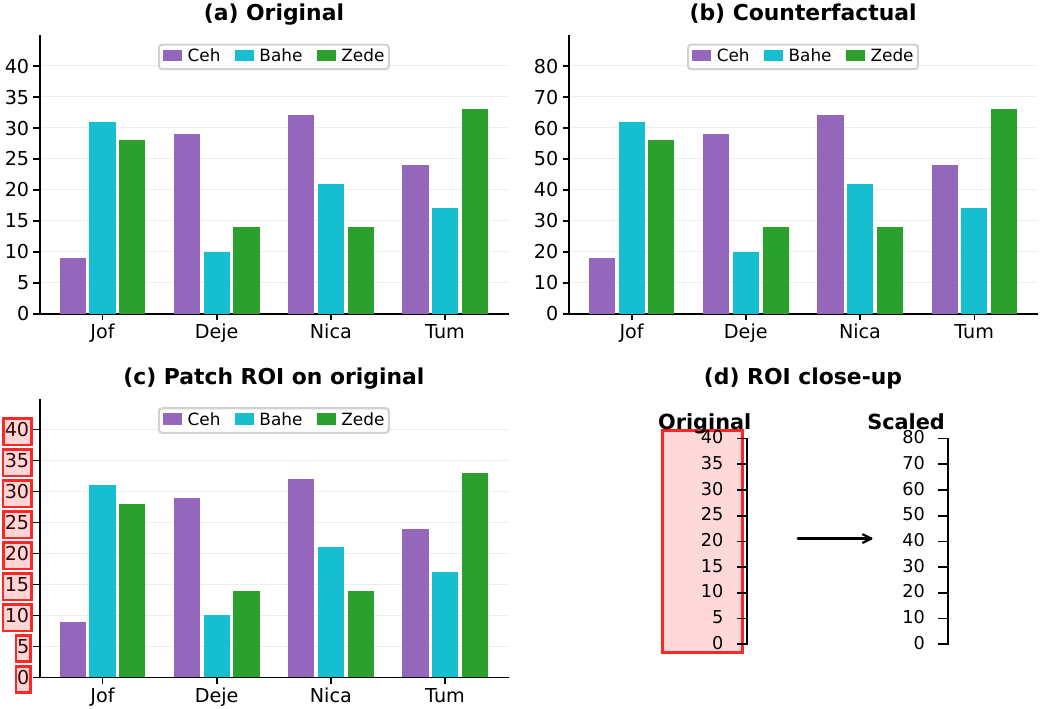}
  \caption{Example of the axis-scale counterfactual. The y-axis scale is rewritten while the patch ROI covers the affected y-axis tick-text region.}
  \label{fig:vertical-bar-axis-scale}
\end{figure*}

\clearpage

\end{document}